\documentclass[journal]{IEEEtran}

\usepackage{cite}
\usepackage{amsmath}
\usepackage{amssymb}
\usepackage{times}
\usepackage{multirow}
\usepackage{color}
\usepackage[switch]{lineno}
\usepackage{graphicx}
\usepackage{subfigure}
\usepackage{url}

\usepackage{algorithm}
\usepackage{algorithmic}
\begin{document}
%\pagewiselinenumbers
\title{Neural Task Planning with \\ And-Or Graph Representations}

\author{Tianshui~Chen, Riquan~Chen, Lin~Nie, Xiaonan~Luo, Xiaobai Liu, and Liang~Lin
\thanks{T. Chen, R. Chen, L. Lin and L. Lin are with the School of Data and Computer Science, Sun Yat-sen University, Guangzhou 510006, China. E-mail: linliang@ieee.org.

X. Luo is with the School of Computer Science and Information Security, Guilin University of Electronic Technology, Guilin 541004, China. E-mail: luoxn@guet.edu.cn.

X. Liu is with the Department of Computer Science, San Diego State University, San Diego, CA, USA. E-mail: xiaobai.liu@sdsu.edu.}}

\maketitle

\begin{abstract}
This paper focuses on semantic task planning, i.e., predicting a sequence of actions toward accomplishing a specific task under a certain scene, which is a new problem in computer vision research. The primary challenges are how to model task-specific knowledge and how to integrate this knowledge into the learning procedure. In this work, we propose training a recurrent long short-term memory (LSTM) network to address this problem, i.e., taking a scene image (including pre-located objects) and the specified task as input and recurrently predicting action sequences. However, training such a network generally requires large numbers of annotated samples to cover the semantic space (e.g., diverse action decomposition and ordering). To overcome this issue, we introduce a knowledge and-or graph (AOG) for task description, which hierarchically represents a task as atomic actions. With this AOG representation, we can produce many valid samples (i.e., action sequences according to common sense) by training another auxiliary LSTM network with a small set of annotated samples. Furthermore, these generated samples (i.e., task-oriented action sequences) effectively facilitate training of the model for semantic task planning. In our experiments, we create a new dataset that contains diverse daily tasks and extensively evaluate the effectiveness of our approach.
\end{abstract}


\IEEEpeerreviewmaketitle

\section{Introduction}
\begin{figure}[!t]
   \centering
   \includegraphics[width=1.0\linewidth]{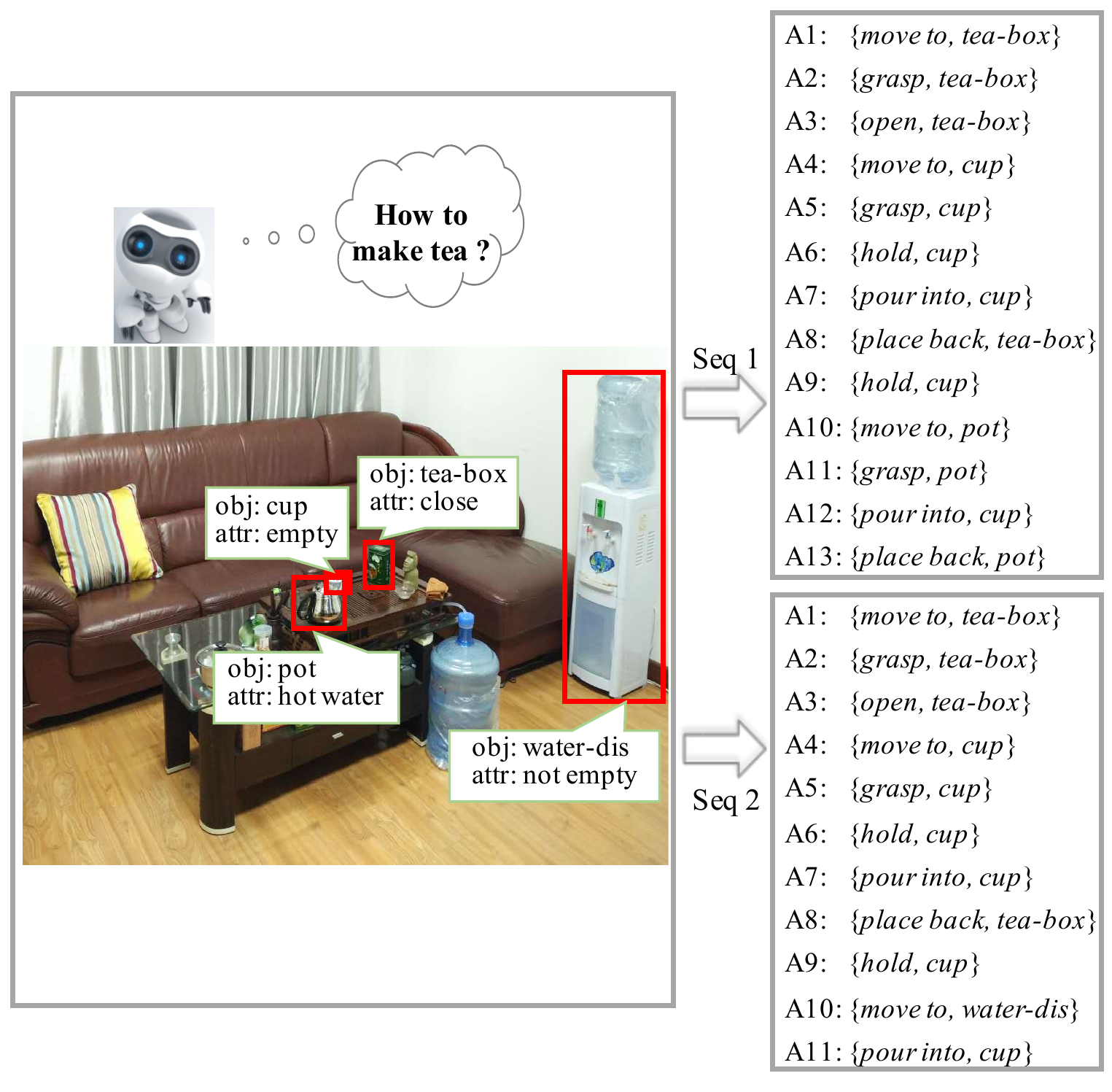}
   \caption{Two alternative action sequences,  inferred according to the joint understanding of the scene image and task semantics, for completing the task ``make tea'' under a given office scene. An agent can achieve this task by successively executing either of the  sequences.}
   \label{fig:task-definition}
\end{figure}

Automatically predicting and executing a sequence of actions given a specific task is an ability that is quite expected for intelligent robots~\cite{thrun2005probabilistic,parker1996alliance}. For example, to complete the task of  ``make tea'' under the scene shown in Figure \ref{fig:task-definition}, an agent needs to plan and successively execute a number of steps, e.g., ``move to the tea box'' and ``grasp the tea box''. In this paper, we aim to train a neural network model to enable this capability, which has rarely been addressed in computer vision research.

We regard the aforementioned problem as semantic task planning, i.e., predicting a sequence of atomic actions toward accomplishing a specific task. Furthermore, we consider an atomic action to be a primitive action operating on an object, denoted by a two-tuple $A=(action, object)$. Therefore, the prediction of action sequences depends on not only the task semantics (i.e., how the task is represented and planned) but also the visual scene image parsing (e.g., recognizing object categories, states and their spatial relations in the scene). Considering the task of a robot pouring a cup of water from a pot, the predicted sequence varies according to the properties of the objects in the scene such as the relative distances among the agent, cup and pot and the state of the cup (empty or not). If the robot is located far from the cup, it should first move close to the cup and then grasp the cup. If the cup is full of water, the robot will have to pour  the water out before filling the cup with water from the pot.  
Since recent advanced deep convolutional neural networks (CNNs) \cite{krizhevsky2012imagenet, zeiler2014visualizing, simonyan2014very, he2016deep} have achieved great successes in object categorization \cite{wang2015large,abdulnabi2015multi,zhao2017diversified,chen2018recurrent,zhou2017multi,chen2018knowledge,chen2018fine} and localization \cite{ren2015faster,li2017attentive,chen2018learning,chen2016disc}, we assume that objects are correctly located and that the initial states of the objects are known in the given scene in this work. However, this problem remains challenging due to the diversity of action decomposition and ordering, long-term dependencies among atomic actions, and large variations in object states and layouts in the scene. 

In this work, we develop a recurrent long short-term memory (LSTM) \cite{hochreiter1997long} network to address the problem of semantic task planning because LSTM networks have been demonstrated to be effective in capturing long-range sequential dependencies, especially for tasks such as machine translation \cite{cho2014learning} and image/video captioning \cite{vinyals2015show,gao2017video}. These approaches generally adopt an encoder-decoder architecture, in which an encoder  first encodes the input data (e.g., an image) into a semantic-aware feature representation and a decoder  then decodes this representation into the target sequence (e.g., a sentence description). In this work, we transform the input image into a vector that contains the information about the object categories and locations and then feed this vector into the LSTM network (named Action-LSTM) with the specified task. This network is capable of generating the action sequence through the encoder-decoder learning.

%We develop a recurrent long short-term memory (LSTM) \cite{hochreiter1997long} network to address the problem of semantic task planning since LSTM models have been demonstrated to be effective in capturing long-range sequential dependencies, particularly for tasks such as machine translation \cite{cho2014learning} and image/video captioning \cite{vinyals2015show,gao2017video}. These approaches generally adopt an encoder-decoder architecture, in which an encoder  first encodes the input data (e.g., an image) into a semantic-aware feature representation and a decoder  then decodes this representation into the target sequence (e.g., a sentence description). In this work, we transform the input image into a vector that contains the information of object categories and locations and feed this vector into the LSTM network (named Action-LSTM) with the specified task, and this network is capable of generating the action sequence through the encoder-decoder learning.  

In general, large numbers of annotated samples are required to train LSTM networks, especially for complex problems such as semantic task planning. To overcome this issue, we present a two-stage training method by employing a knowledge  and-or graph (AOG) representation \cite{lin2009semantic, xiong2016robot,li2017joint}. First, we define the AOG for task description, which hierarchically decomposes a task into atomic actions according to their temporal dependencies. In this semantic representation, an and-node represents the chronological decomposition of a task (or sub-task), an or-node represents the alternative ways to complete the certain task (or sub-task), and leaf nodes represent the pre-defined atomic actions. The AOG can thus contain all possible action sequences for each task by embodying the expressiveness of grammars. Specifically, given a scene image, a specific action sequence can be generated by selecting the sub-branches at all of the or-nodes in a depth-first search (DFS) manner. Second, we train an auxiliary LSTM network (named AOG-LSTM) to predict the selection at the or-nodes in the AOG and thus produce a large number of new valid samples (i.e., task-oriented action sequences) that can be used for training the Action-LSTM network. Notably, training the AOG-LSTM network requires only a few manually annotated samples (i.e., scene images and the corresponding action sequences) because making a selection in the context of task-specific knowledge (represented by the AOG) is seldom ambiguous.

Note that a preliminary version of this work has been presented at a conference \cite{lin2017knowledge}. In this paper, we inherit the idea of integrating task-specific knowledge via a two-stage training method, and we extend the initial version from several aspects to strengthen our method. First, we extend the benchmark to involve more tasks and include more diverse scenarios. Moreover, because the automatically augmented set includes some difficult samples with uncertain and even incorrect labels, we further incorporate curriculum learning \cite{bengio2009curriculum,khan2011humans} to address this issue by starting the training with only easy samples and then gradually extending to more difficult samples. Finally, more detailed comparisons and analyses are conducted to demonstrate the effectiveness of our proposed model and to verify the contribution of each component.

The main contributions of this paper are two-fold. First, we present a new problem called semantic task planning and create a benchmark (that includes 15 daily tasks and 1,284 scene images). Second, we propose a general approach for incorporating complex semantics into the recurrent neural network (RNN) learning, which can be generalized to various high-level intelligent applications.

The remainder of this paper is organized as follows. Section II provides a review of the most-related works. Section III presents a brief overview of the proposed method. We then introduce the AOG-LSTM and Action-LSTM modules in detail in Sections IV and V, respectively, with thorough analyses of the network architectures, training and inference processes of these two modules. Extensive experimental results, comparisons and analyses are presented in Section VI. Finally, Section VII  concludes this paper.

\section{Related work}
We review the related works following three main research streams: task planning, action recognition and prediction, and recurrent sequence prediction. 

\subsection{Task planning}
In the literature, task planning (also referred to as symbolic planning~\cite{sung2013learning}) has traditionally been formalized as  deduction \cite{allen1991planning,smith1999temporal} or satisfiability \cite{kautz1992planning, rintanen2006planning} problems for  long periods. Sacerdoti et al. \cite{sacerdoti1974planning} introduced hierarchical planning, which first performed planning in an abstract manner and then generated fine-level details. Yang et al. \cite{yang2007learning} utilized the standard Planning Domain Definition Language (PDDL) representation for actions and developed an action-related modeling system to learn an action model from a set of observed successful plans. Some work also combined symbolic planning with motion planning \cite{kambhampati1991combining}. Cambon et al. \cite{cambon2009hybrid} regarded symbolic planning as a constraint and proposed a heuristic function for motion planning. Plaku et al. \cite{plaku2010sampling} extended the work and planned using geometric and differential constraints. Wolfe et al. \cite{wolfe2010combined} proposed a hierarchical task and motion planning algorithm based on hierarchical transition networks. Although those algorithms performed quite well in controlled environments, they required encoding every precondition for each operation or domain knowledge, and they can hardly be generalized to unconstrained environments with large variances \cite{sung2013learning}. Recently, Sung et al. \cite{sung2013learning, sung2014synthesizing} represented an environment with a set of attributes and proposed using a Markov random field (MRF) \cite{li1994markov} to learn the sequences of controllers to complete the given tasks. Xiong et al. \cite{xiong2016robot} developed a stochastic graph-based framework, which incorporated spatial, temporal and causal knowledge, for a robot to understand tasks, and they successfully applied this framework to a cloth-folding task.

\begin{figure*}[!t]
   \centering
   \includegraphics[width=0.95\linewidth]{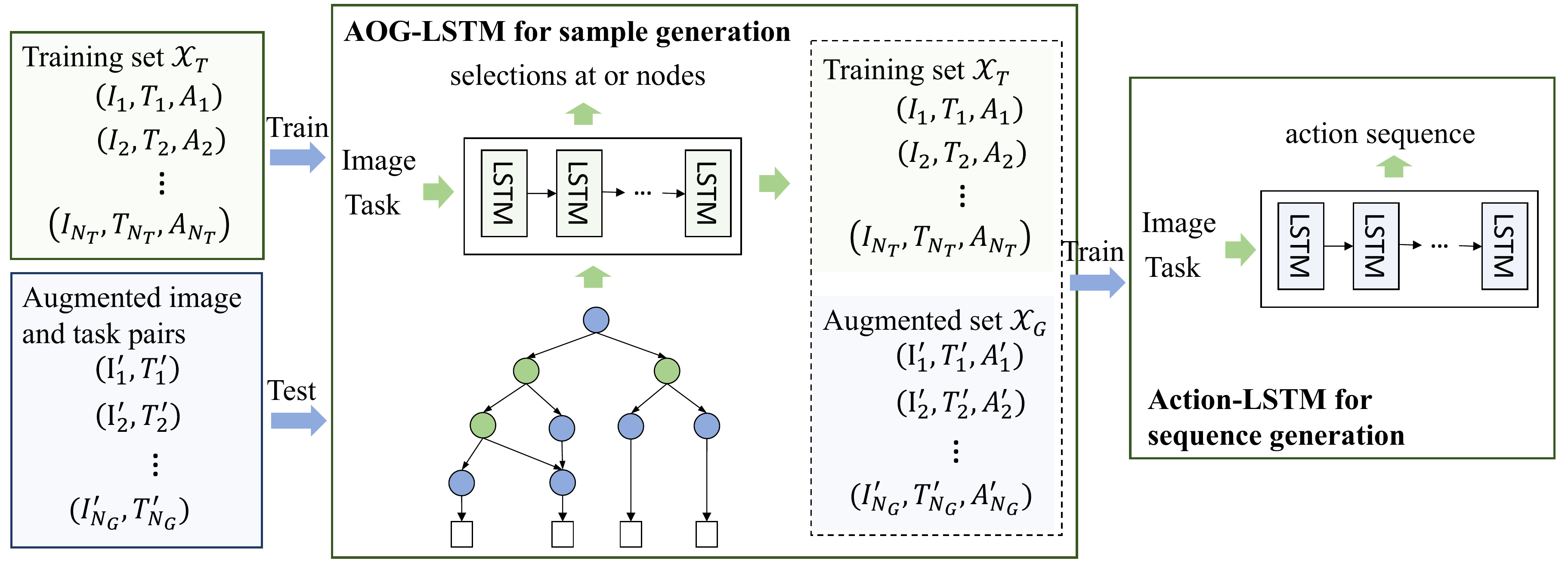}
   \caption{An overall introduction of the proposed method. The AOG-LSTM network is trained using the samples from the small training set, and it can be used to generate a relatively large augmented set. The augmented set, together with the training set, is used to train the Action-LSTM network, which can directly predict the action sequence to complete a given task under a certain scene.}
   \label{fig:framework}
\end{figure*}

Some other works also manually defined the controller sequences for completing certain tasks, including baking cookies \cite{bollini2011bakebot}, making pancakes \cite{beetz2011robotic}, and folding laundry \cite{miller2012geometric}. In these works, the controller sequences were selected from several predefined sequences or retrieved from a website. For example, in the work \cite{beetz2011robotic}, the robot retrieved instructions for making pancakes from the Internet to generate the controller sequence. Although they often achieve correct sequences in controlled environments, these methods cannot scale up to a large number of tasks in unconstrained household environments because each task requires defining complicated rules for the controller sequence to adapt to various environments.

Recently, a series of works \cite{zhu2017visual,gupta2017cognitive} developed learning-based planners for semantic visual planning tasks. Gupta et al. \cite{gupta2017cognitive} proposed a  cognitive mapper and planner for visual navigation, therein constructing a top-down belief map of the world and applying a differentiable neural network planner to produce the next action at each time step. Zhu et al. \cite{zhu2017visual} formulated visual semantic planning as a policy learning problem and proposed an interaction-centric solution that offered crucial transferability properties for semantic action sequence prediction. In addition to visual semantic planning, the deep neural networks were also adopted to address the tasks of robotic control planning \cite{pascanu2017learning,agrawal2016learning,finn2017deep}. For example, Agrawal et al. \cite{agrawal2016learning} developed a joint forward and inverse models for real-world robotic manipulation tasks. The inverse model provided supervision to construct informative visual features, which the forward model then predicted and in turn regularized the feature space for the inverse model. Pascanu et al. \cite{pascanu2017learning} introduced an imagination-based planner that could learn to construct, evaluate, and execute plans.

\begin{figure*}[!t]
\centering
\subfigure[]{
\includegraphics[width=0.48\linewidth]{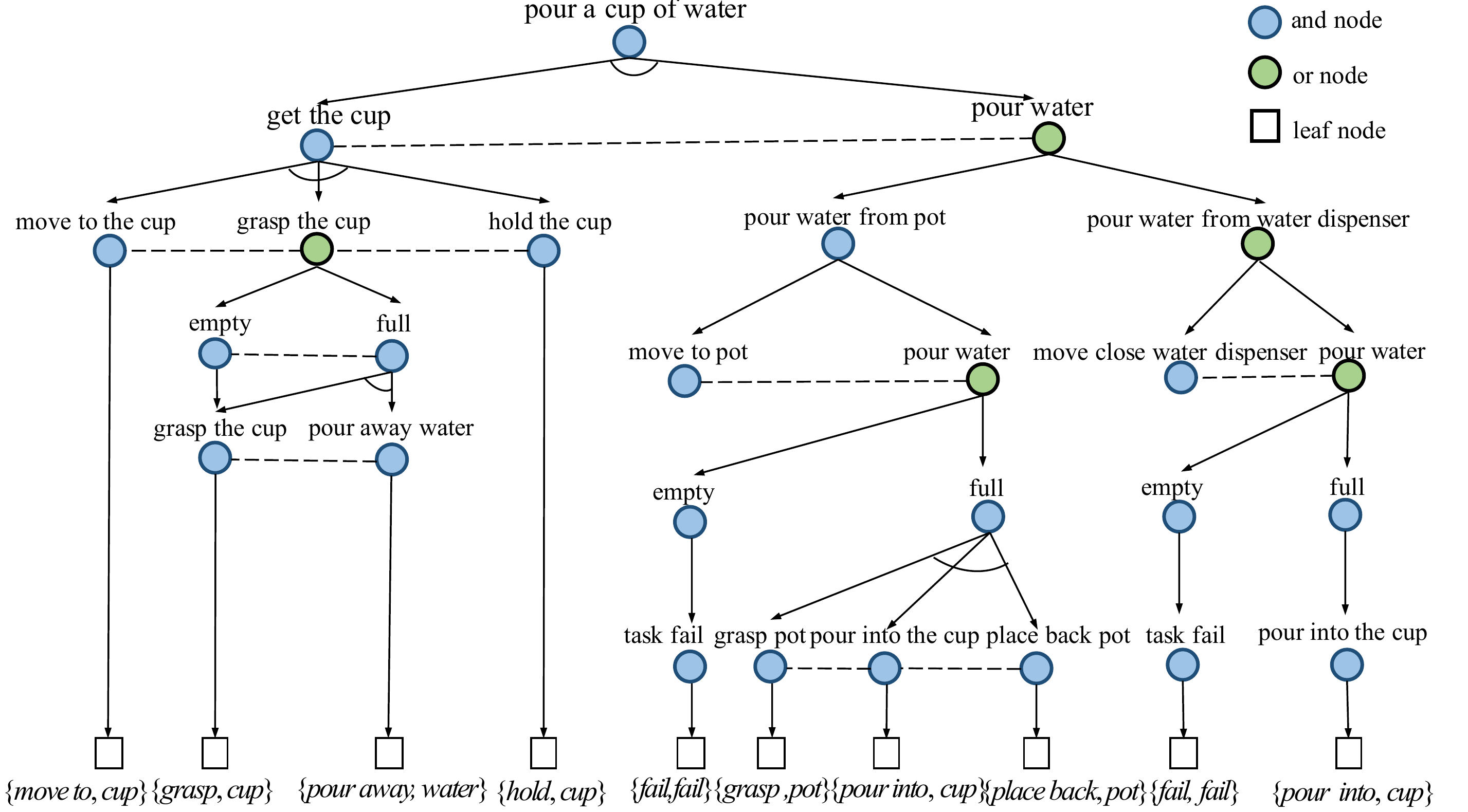}}
\subfigure[]{
\includegraphics[width=0.50\linewidth]{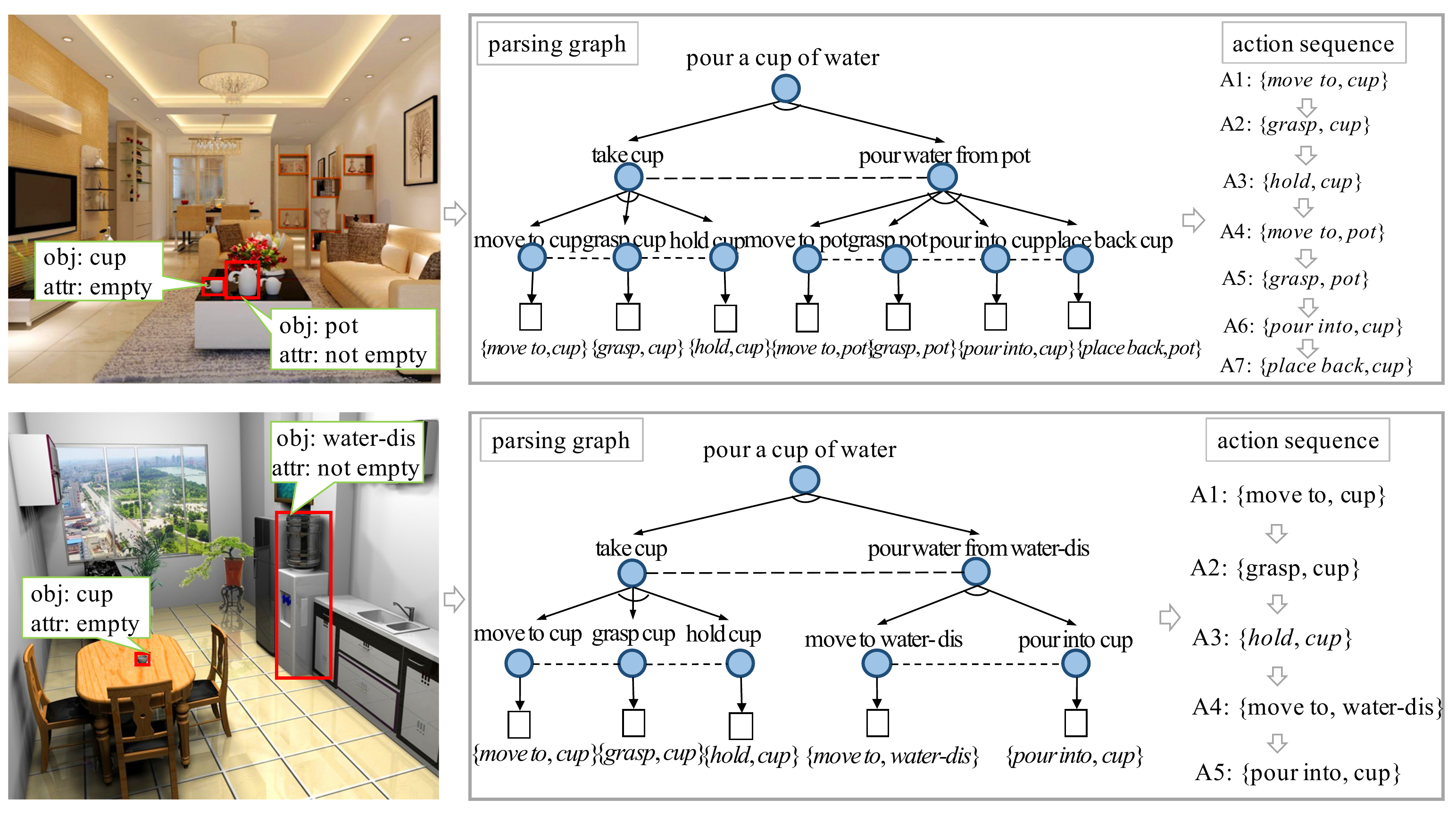}}
\caption{An example of a knowledge and-or graph for describing the task ``pour a cup of water''  shown in (a) and two parsing graphs and their corresponding action sequences under two specific scenes shown in (b).}
\label{fig:aog}
\end{figure*}

\subsection{Action recognition and prediction}
The problem studied in this paper is also related to action recognition and prediction, where the former attempts to recognize action categories performed by persons from a fully observed video/image \cite{wu2014human,xu2017hierarchical,samanta2014space,guo2016tensor,lin2016deep} and the latter targets predicting an action that humans are likely to perform in the future within given scenarios  \cite{kong2014discriminative,lan2014hierarchical,wang2017context}. Note that our work differs from the aforementioned methods in the goal of the problem, which is to automatically infer potential action sequences that can be used to complete the task at hand in the specific environment. Note that action recognition and prediction can be beneficial to our work because they provide
a better understanding of the interaction between humans and the environment for the robots for task planning.

\subsection{Recurrent sequence prediction} 
RNNs \cite{connor1994recurrent}, particularly LSTM \cite{hochreiter1997long}, were developed for modeling long-term temporal dependencies. Recently, RNNs have been extensively applied to various sequence prediction tasks, including natural language generation \cite{wen2015semantically, rush2015neural,tang2016context}, neural machine translation \cite{cho2014learning, bahdanau2014neural,cho2014properties,wu2016google}, and image and video captioning \cite{vinyals2015show, yao2015describing,you2016image,pan2016hierarchical,gao2017video}. These works adopted a similar encoder-decoder architecture for solving sequence prediction. Tang et al. \cite{tang2016context} encoded the contexts into a continuous semantic representation and then decoded the semantic representation into context-aware text sequences using RNNs. Cho et al. \cite{cho2014learning} mapped a free-form source language sentence into the target language by utilizing the encoder-decoder recurrent network. Vinyals et al. \cite{vinyals2015show} applied a similar pipeline for image captioning, therein employing a CNN as the encoder to extract image features and an LSTM network as the decoder to generate the descriptive sentence. Pan et al. \cite{pan2016hierarchical} further adapted this pipeline to video caption generation by developing a hierarchical recurrent neural encoder that is capable of efficiently exploiting the video temporal structure in a longer range to extract video representations.

\section{Overview}
In this section, we give an overall introduction to the proposed method. First, we represent the possible action sequences for each task with an AOG. Based on this AOG, a parsing graph, which corresponds to a specific action sequence, can be generated by selecting the sub-branches at all the or-nodes searched in a DFS manner given a scene image. An LSTM (namely, AOG-LSTM) is learned to make predictions at these or-nodes given a large number of new scene image and automatically produce a relatively large number of valid samples. Finally, these automatically generated samples are used to train another LSTM (namely, Action-LSTM) that directly predicts the action sequence to complete a given task under a certain scene. An overall illustration is presented in Figure \ref{fig:framework}.

%\vspace{6pt}

%In this section, we first present an overall introduction of the proposed method. We represent the possible action sequences for each task with an AOG, from which a parsing graph corresponding to a specific action sequence will be generated by selecting the sub-branches at all the or-nodes searched in a DFS manner given a new scene image. An LSTM (namely AOG-LSTM), which is trained with a few samples, is adopted to make predictions at these or-nodes, and it can further be used to generate a large number of new valid samples. Finally, another LSTM (namely Action-LSTM), which is trained with both manually-annotated and automatically-generated samples, are utilized to directly predict the action sequence to complete a given task under a certain scene. An overall description of the proposed method is illustrated in Figure \ref{fig:framework}.

\section{Semantic Task Representation}

\subsection{Atomic action definition}
An atomic action refers to a primitive action operating on an associated object, and it is denoted as a two-tuple set $A=(action, object)$. To ensure that the learned model can generalize across different tasks, the primitive action and associated object should satisfy two properties \cite{sung2014synthesizing}: 1) each primitive action should specialize an atomic operation, such as open, grasp or move to, and 2) the primitive actions and associated objects should not be specific to one task. With these role-specific and generalizable settings, large numbers of high-level tasks can be completed using the atomic actions defined on a small set of primitive actions and associated objects. In this work, $B_a=$12 primitive actions and $B_o=$12 objects are involved, as described in Section \ref{sec:dataset}.

\subsection{Knowledge and-or graph}
\label{sec:taog}
The AOG is defined as a 4-tuple set $\mathcal{G}=\{S, V_N, V_T, P\}$, where $S$ is the root node denoting a task. The non-terminal node set $V_N$ contains both and-nodes and or-nodes. An and-node represents the decomposition of a task to its sub-tasks in chronological order. An or-node is a switch, deciding which alternative sub-task to select. Each or-node has a probability distribution $\mathbf{p}_t$ (the $t$-th element of $P$) over its child nodes, and the decision is made based on this distribution. $V_T$ is the set of terminal nodes. In our definition of AOG, the non-terminal nodes refer to the sub-tasks and atomic actions, and the terminal nodes associate a batch of atomic actions. In this work, we manually define the structure of the AOG for each task.

According to this representation, the task ``pour a cup of water'' can be represented as the AOG shown in Figure \ref{fig:aog}(a). The root node denotes the task, where is first decomposed into two sub-tasks, i.e., ``get the cup'' and ``pour water'', under the temporal constraint. The ``get the cup'' node is an and-node and can be further decomposed into ``move to the cup'', ``take the cup'' and ``hold the cup'' in chronological order. The ``pour water'' node is an or-node, and it has two alternative sub-branches, i.e., ``pour water from the water dispenser'' and ``pour water from the pot''. Finally, all the atomic actions are treated as the primitive actions and associated objects, which are represented by the terminal nodes. In this way, the knowledge AOG contains all possible action sequences of the corresponding task in a syntactic manner.

\begin{figure}[!t]
   \centering
   \includegraphics[width=1.0\linewidth]{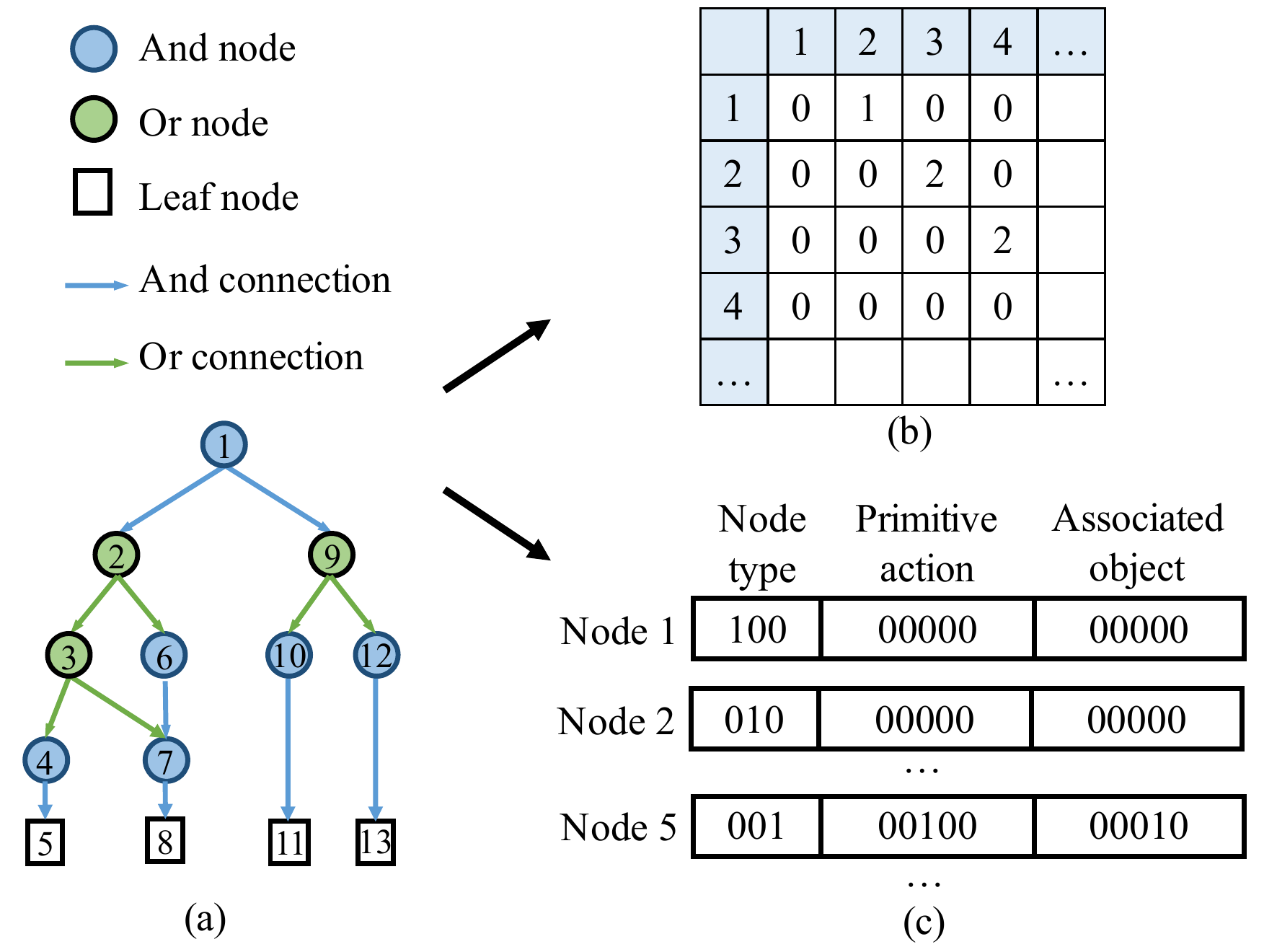}
   \caption{Illustration of the AOG encoding process. The nodes in the AOG are first numbered (a). Then, an adjacency matrix is employed to encode the structure of the graph, with a value of 1 for an and connection, 2 for an or connection and 0 otherwise (b). The content of each node contains a one-hot vector denoting the node type, two one-hot vectors representing the primitive action and the associated object of the associated atomic action for the leaf node, and two zero vectors for the and- and or-nodes.}
   \label{fig:aog-encoding}
\end{figure}

\subsection{Sample generation with and-or graph}
\label{sec:sagwaog}
In addition to capturing the task semantics, the AOG representation enables the generation of a large number of valid samples (i.e., action sequences extracted from the AOG), which are important for the RNN learning process. According to the definition of the AOG, a parsing graph, which corresponds to a specific action sequence (e.g., Figure \ref{fig:aog}(b)), will be generated by selecting the sub-branches for all the or-nodes searched in a DFS manner given a new scene image. Since explicit temporal dependencies exist among these or-nodes, we can recurrently activate these selections using an LSTM network, i.e., AOG-LSTM.

\noindent\textbf{AOG encoding. }Before discussing the AOG-LSTM, we first introduce how to encode the AOG into a feature vector because this is a crucial process for integrating the AOG representation into the LSTM.  Concretely, the AOG features should contain both graph structure and node content information, and the encoding process consists of three steps, as illustrated in Figure \ref{fig:aog-encoding}. First, we number all the nodes in the AOG, as shown in Figure \ref{fig:aog-encoding}(a). Second, an adjacency matrix \cite{harary1962determinant} is utilized to encode the graph structure that depicts whether and how two nodes are connected. Consistent with the situation whereby and- and or-nodes exist in the AOG and they represent completely different meanings, we define an and connection to signify a connection of an and-node to its child and an or connection to signify that of an or-node to its child. Suppose that the adjacency matrix is $M$, with $M_{ij}$ being the value of row $i$ and column $j$; $M_{ij}$ is set to 1 for the and connection of the and-node $i$ to its child $j$, 2 for the or connection and 0 otherwise (see Figure \ref{fig:aog-encoding}(b)). Third, we extract the features for each node to encode its node type and related atomic action information. There are three types of nodes in the AOG, i.e., and, or and leaf nodes, and we utilize one-hot vectors to represent these nodes. As indicated in the definition of the AOG, a leaf node is connected to a specific atomic action, and we employ two one-hot vectors to represent the primitive action and associated object of the atomic action. Then, we append the two vectors after the node type vector to obtain the features of this node. In contrast, the and- and or-nodes do not connect to the specific atomic action directly; thus, we simply pad zeros after the node type vector (see Figure \ref{fig:aog-encoding}(c)). Finally, the adjacency matrices are re-arranged to a vector, and the vector is concatenated with all the node features to achieve the final representation of the AOG.

\begin{figure}[!t]
   \centering
   \includegraphics[width=1.0\linewidth]{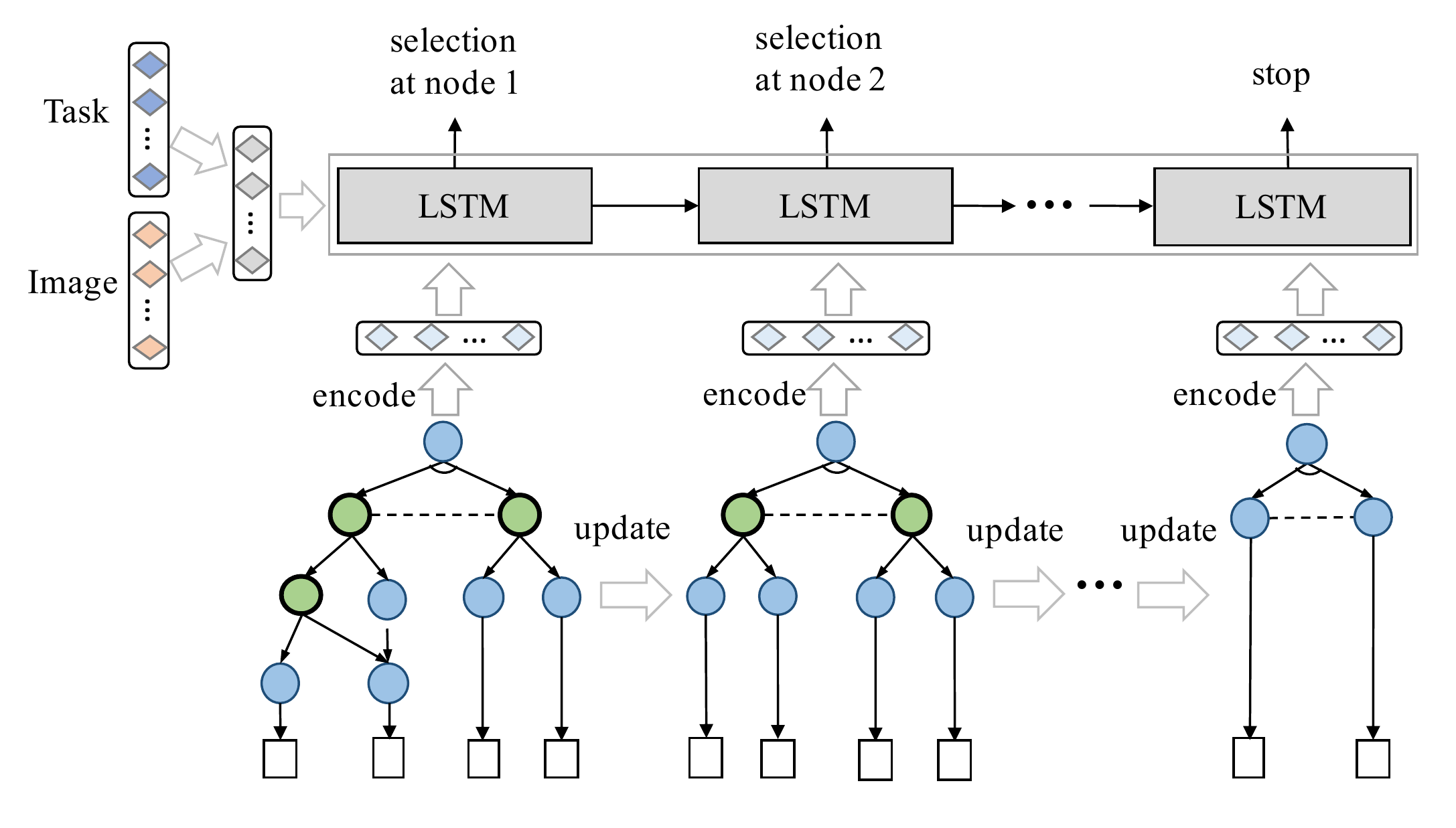}
   \caption{The AOG-LSTM architecture for selecting the sub-branches at all of the or-nodes in a knowledge and-or graph.}
   \label{fig:aog-lstm}
\end{figure}

According to the AOG definition, we search the or-nodes based on the depth-first and from left-to-right manner. As illustrated in Figure \ref{fig:aog-lstm}, our model first extracts the features of the given scene image and the task, and it maps them to a feature vector, which serves as the initial hidden state of the AOG-LSTM. The model then encodes the initial AOG as a feature vector, which is fed into the AOG-LSTM to predict the sub-branch selection of the first or-node. Meanwhile, the AOG is updated by pruning the unselected sub-branches. Note that the AOG is updated based on the annotated and predicted selections during the training and test stages, respectively. Based on the updated AOG, the same process is conducted to predict the selection of the second or-node. This process is repeated until all or-nodes have been visited, and a parsing graph is then constructed. We denote the image and task features as $\mathbf{f}^I$ and $\mathbf{f}^T$, respectively, and we denote the AOG features at time step $t$ as $\mathbf{f}_t^{AOG}$. The prediction at time step $t$ can be expressed as follows:

\begin{equation}
   \begin{split}
   &\mathbf{f}_{IT}=[\mathrm{relu}(\mathbf{W}_{fI}\mathbf{f}^I), \mathrm{relu}(\mathbf{W}_{fT}\mathbf{f}^T)] \\
   &\mathbf{c}_0 =  \mathbf{0}; \ \mathbf{h}_0=\mathbf{W}_{hf}\mathbf{f}_{IT} \\
   &[\mathbf{h}_t, \mathbf{c}_t] = \mathrm{LSTM}(\mathbf{f}^{AOG}_t, \mathbf{h}_{t-1}, \mathbf{c}_{t-1}) \\
   &\mathbf{p}_t = \mathrm{softmax}(\mathbf{W}_{hp}\mathbf{h}_t + \mathbf{b}_p)
   \end{split}
   \label{eqn:AOG-LSTM}
\end{equation}
where $\mathrm{relu}$ is the rectified linear unit (ReLU) function \cite{nair2010rectified} and $\mathbf{p}_t$ is the probability distribution over all child branches of the $t$-th or-node, where the branch with the maximum value is selected. 
$\mathbf{W}_{fI}$, $\mathbf{W}_{fT}$, $\mathbf{W}_{hf}$, and $\mathbf{W}_{hp}$ are the parameter matrices, and $\mathbf{b}_{p}$ is the corresponding bias vector. $\mathbf{f}^I$ are the image features containing the information of the class labels, initial states, and locations of the objects in image $I$. More concretely, suppose that there are $B_o$ categories of objects and $k$ attributes. For each object in an given image, we utilize a $B_o$-dimension one-hot vector to denote its class label information, $k$ vectors to denote the initial states of the $k$ attributes, and a 4-dimension vector to denote the bounding box of the object region. Thus, each object can be represented by a fixed dimension feature vector, and the feature vectors of all objects are concatenated to obtain an image feature vector. However, the number of objects varies in different images, leading to image feature vectors with different dimensions. To address this issue, we first extract the features for the image with the maximum number of objects, and we apply zero padding to each feature vector so that each vector has the same dimensions as the feature vector of the image with the maximum number of objects. $\mathbf{f}^T$ is a one-hot vector denoting a specific task. $\mathbf{f}^I$ and $\mathbf{f}^T$ are first processed using two separated fully connected layers followed by the ReLU function to generate two 256-D feature vectors. The initial memory cell $\mathbf{c}_0$ is set as a zero vector. The two vectors are then concatenated and mapped to a 256-D feature vector using a fully connected layer, which serves as the initial hidden state of the LSTM. $\mathbf{f}^{AOG}_t$ are the AOG feature vectors at time step $t$, which are also pre-processed to a 256-D feature vector via a fully connected layer and then fed to the AOG-LSTM. The size of the hidden layer of the LSTM is 256 neurons.

\noindent\textbf{AOG-LSTM training. }Making a selection at the or-nodes is less ambiguous because the AOG representation effectively regularizes the semantic space. Thus, we can train the AOG-LSTM using only a small number of annotated samples. Specifically, we collect a small set of samples annotated with the selections of all or-nodes given a scene image for each task, i.e., $\mathcal{X}_T=\{I_n, T_n, \mathbf{s}_n\}_{n=1}^{N_T}$, in which $I_n$ and $T_n$ are the $n$-th given image and task, respectively, and $\mathbf{s}_n=\{s_{n1}, s_{n2}, \dots, s_{n{K_n}}\}$ is a set whereby $s_{nb}$ denotes the selection for the $t$-th or-node and $K_n$ is the number of  or-nodes. $N_T$ is the number of annotated samples in $\mathcal{X}_T$. Given the predicted probability $\mathbf{p}_{nt}=\{p_{nt1}, p_{nt2}, \dots, p_{ntB}\}$ for the $t$-th or-node, we define the objective function as the sum of the negative log-likelihood of correct selections over the entire training set, which is formulated as
\begin{equation}
   \mathcal{L}_{aog}=-\sum_{n=1}^{N_T}\sum_{t=1}^{K_n}\sum_{b=0}^{B}{\mathbf{1}(s_{nb}=b)\log{p_{ntb}}},
   \label{eqn:aog_loss}
\end{equation}
where $\mathbf{1}(\cdot)$ is an indicator function whose value is 1 when the expression is true and 0 otherwise and $B$ is the number of sub-branches. In our experiment, because the maximum number of sub-branches is 3, we simply set $B$ to 3.

\noindent\textbf{Sample generation. }Once the AOG-LSTM is trained,  we use it to predict the sub-branch selections for all the or-nodes in the AOG given different scene images and generate the corresponding action sequences. In this way, a relatively large set of $\mathcal{X}_G=\{I_n,T_n,\mathcal{A}_n\}_{n=1}^{N_G}$ is obtained, where $I_n$, $T_n$, and $\mathcal{A}_n$ represent the image, task and predicted sequence for the $n$-th sample, respectively, and $N_G$ is the number of generated samples. More importantly, it can also generate samples of unseen tasks using an identical process in which the AOG structures for the new tasks are also manually defined. These newly generated samples effectively alleviate the problem of manually annotating large numbers of samples in practice.

\section{Recurrent action prediction}
\label{sec:rap}
We formulate the problem of semantic task planning in the form of the probability estimation $p(A_1, A_2, ..., A_n|I,T)$, where $I$ and $T$ are the given scene image and the task, respectively, and $\{A_1, A_2, ..., A_n\}$ denotes the predicted sequence. Based on the chain rule, the probability can be recursively decomposed as follows:
\begin{align}
   p(A_1, A_2, ..., A_n|I, T)=\prod_{t=1}^np(A_t|I, T, \mathcal{A}_{t-1}),
   \label{eqn:chainrule}
\end{align}
where $\mathcal{A}_{t-1}$ denotes $\{A_1, A_2, ..., A_{t-1}\}$ for convenience of illustration. The atomic action is defined as $A_i=(a_i, o_i)$. Since an atomic action is composed of a primitive action and an associated object, there are large numbers of atomic actions that have few samples because action-object co-occurrence is infrequent in the training samples. Thus, a fundamental problem in atomic action prediction is learning from very few samples. Fortunately, although the atomic action might occur rarely in the samples, its primitive action and associated object independently appear quite frequently. Thus, in this work, we simplify the model by assuming independence between the primitive actions and the associated objects and predict them separately \cite{zhu2014reasoning}. The probability can be expressed as follows:
\begin{align}
   p(A_t|I, T, \mathcal{A}_{t-1})=p(a_t|I, T, \mathcal{A}_{t-1})p(o_t|I, T, \mathcal{A}_{t-1}).
   \label{eqn:decomposition}
\end{align}

\begin{figure}[!t]
   \centering
   \includegraphics[width=1.0\linewidth]{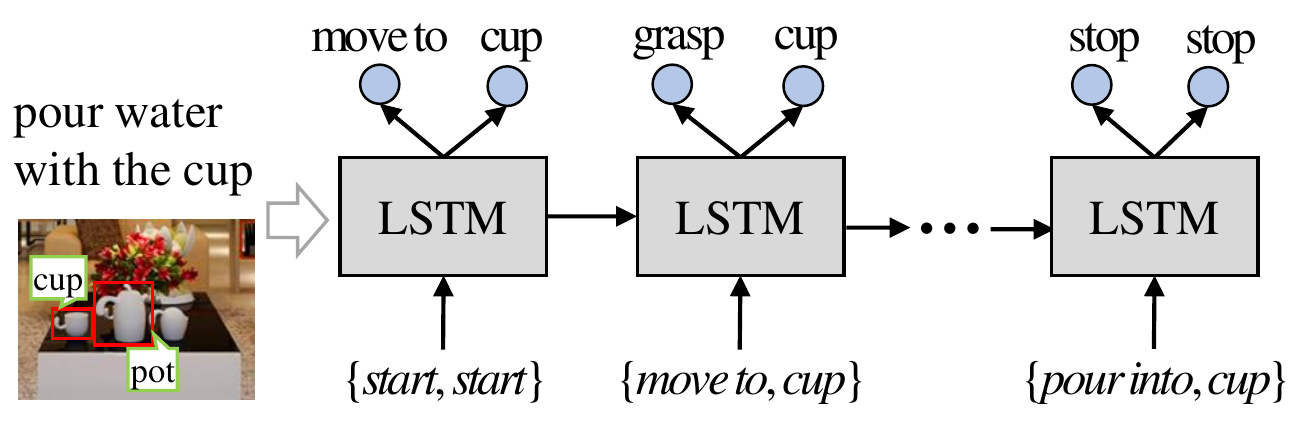} % requires the graphicx package
   \caption{The Action-LSTM architecture for predicting the atomic action sequence given a specific task.}
   \label{fig:s-lstm}
\end{figure}

Here, we develop the Action-LSTM network to model the probability distribution, i.e., equation (\ref{eqn:chainrule}).  Specifically, the Action-LSTM network first applies a process similar to that of AOG-LSTM to extract the features of the task and image, which is also used to initialize the hidden state of the LSTM. At each time step $t$, two softmax layers are utilized to predict the probability distributions $\mathbf{p}(a_t)$ over all primitive actions and $\mathbf{p}(o_t)$ over all associated objects. The conditions on the previous $t-1$ actions can be expressed by the hidden state $h_{t-1}$ and memory cell $c_{t-1}$. The action prediction at time step $t$ can be computed as follows:
\begin{equation}
   \begin{split}
    &\mathbf{f}_{IT}=[\mathrm{relu}(\mathbf{W}_{fI}\mathbf{f}^I), \mathrm{relu}(\mathbf{W}_{fT}\mathbf{f}^T)] \\
   &\mathbf{c}_0 =  \mathbf{0}; \ \mathbf{h}_0=\mathbf{W}_{hf}\mathbf{f}_{IT}\\
   &[\mathbf{h}_t, \mathbf{c}_t] = \mathrm{LSTM}(\mathbf{f}^A_t, \mathbf{h}_{t-1}, \mathbf{c}_{t-1}) \\
   &\mathbf{p}(a_t) = \mathrm{softmax}(\mathbf{W}_{ah}\mathbf{h}_t+\mathbf{b}_a) \\
   &\mathbf{p}(o_t) = \mathrm{softmax}(\mathbf{W}_{oh}\mathbf{h}_t+\mathbf{b}_o)
   \end{split}
   \label{eqn:Action-LSTM}
\end{equation}
where $\mathrm{relu}$ is the ReLU function. $\mathbf{W}_{fI}$, $\mathbf{W}_{fT}$, $\mathbf{W}_{hf}$, $\mathbf{W}_{ah}$, and $\mathbf{W}_{oh}$ are the parameter matrices, and $\mathbf{b}_a$ and $\mathbf{b}_o$ are the corresponding bias vectors. Figure \ref{fig:s-lstm} presents an illustration of the Action-LSTM network. $\mathbf{f}^I$ and $\mathbf{f}^T$ carry exactly the same information as explained for the AOG-LSTM, and they are pre-processed using an identical process except that the last fully connected layer has 512 output neurons and thus generates a 512-D feature vector. Additionally, this feature vector is considered to be the initial hidden state of the Action-LSTM.  $\mathbf{f}^A_t$ is a feature vector that encodes the input atomic action at time step $t$, which is concatenated by two one-hot vectors denoting its primitive action and associated object. Note that the atomic action is the ground truth and predicted atomic action of the previous time step during the training and test stages, respectively. Because directly predicting the atomic action is considerably more complicated, we set the size of the hidden layer of the LSTM to 512 neurons.

\noindent\textbf{Action-LSTM training. }
In the training stage, we leverage the entire training set, including the manually annotated samples and the automatically generated samples, to optimize the network. However, some difficult samples with uncertain or even incorrect labels exist, and these samples may severely impact the model convergence and lead to inferior results. Meanwhile, skipping too many difficult training samples leads to a risk of overfitting on the small set of easy samples, resulting in a poor generalization ability to unseen testing data. To strike a better balance, we employ a curriculum learning algorithm \cite{bengio2009curriculum,khan2011humans} that starts the training process using the most reliable samples and then gradually increases the sample difficulty. 

To determine the difficulty of a particular sample, we consider the uncertainty of making selections at the or-nodes during the sample generation stage. Concretely, a probability distribution is predicted when performing selections at an or-node, and the entropy of this distribution well measures the uncertainty \cite{bialynicki1975uncertainty,jaynes1957information}. Thus, we define the uncertainty of a sample by averaging the entropies of the predicted distributions over all the or-nodes. In this way, a sample having a higher uncertainty means that it is more difficult. We create a curriculum of $X_G$ by sorting the samples according to their uncertainty values and set a threshold $\tau$ to exclude the samples with uncertainty values that are higher than $\tau$. The curriculum is updated by decreasing $\tau$ to include more difficult samples during the training stage. 

Formally, we are given the manually annotated and automatically generated sets, i.e., $\mathcal{X}_T=\{I_n,T_n,\mathcal{A}_n\}_{n=1}^{N_T}$ and  $\mathcal{X}_G=\{I_n,T_n,\mathcal{A}_n\}_{n=1}^{N_G}$, where $I_n$ and $T_n$ are the given image and task of the $n$-th sample, respectively, and $\mathcal{A}_n=\{A_{n1}, A_{n2}, ..., A_{n{W_n}}\}$ is the atomic action sequence, with $W_n$ denoting the number of atomic actions. $A_{nt}=\{a_{nt}, o_{nt}\}$ is the $t$-th atomic action, with $a_{nt}$ and $o_{nt}$ denoting its primitive action and associated object, respectively. Similarly, we define the objective function as the sum of the negative log-likelihood of correct sequences over all the samples in the manually annotated set and the selected samples in the automatically generated set. Given the predicted distribution of the primitive action $\mathbf{p}_n(a_t)=\{p_{n1}(a_t), p_{n2}(a_t), \dots, p_{n{B_a}}(a_t)\}$ and associated object $\mathbf{p}_n(o_t)=\{p_{n1}(o_t), p_{n2}(o_t), \dots, p_{n{B_o}}(o_t)\}$ for the $t$-th step, the objective function can be expressed as

\begin{equation}
   \mathcal{L}_{action}=-\sum_{n=1}^{N_T}\sum_{t=1}^{W_n}\ell_{nt}-\sum_{n'=1}^{N_G}\sum_{t'=1}^{W_{n'}}\mathbf{1}(H_{n'}<\tau)\ell_{{n'}{t'}},
   \label{eqn:action_loss}
\end{equation}
where
\begin{equation}
  \ell_{nt}=\sum_{j=0}^{B_a}{\mathbf{1}(a_{nt}=j)\log{p_{nj}}(a_t)}+\sum_{j=0}^{B_o}{\mathbf{1}(o_{nt}=j)\log{p_{nj}}(o_t)}.
   \label{eqn:action_loss1}
\end{equation}
In these equations, $B_a$ and $B_o$ are the numbers of involved primitive actions and associated objects, $H_{n'}$ is the uncertainty of sample $n'$, and $\mathbf{1}(\cdot)$ is also an indicator function whose value is 1 if the expression is true and 0 otherwise. Note that we directly use all the samples in $\mathcal{X}_T$ because these samples are manually annotated and can effectively avoid samples with uncertain or incorrect labels. 

%For the $i$-th sample, the sample uncertainty $H^i$ can be computed by 
%\begin{align}
%   H^i=\sum_{t=1}^{K_i}\sum_{j=0}^{B_{it}}{p_{tj}^{i}\log{p_{tj}^{i}}}.
%   \label{eqn:sample_uncertainty}
%\end{align}
%where $p_{tj}^{i}$ is the probability of the $j$-th sub-branch at the $t$-th or-node, $B_{it}$ is the number of sub-branches of the $t$-th or-node, $K_i$ is the number of the or-nodes.

\noindent\textbf{Sequence prediction. }Once the Action-LSTM is trained, it is utilized to recurrently predict the atomic action sequence conditioning on the given task and input scene image. Concretely, the Action-LSTM takes a special atomic action $(start, start)$ as input to predict the probability distributions of the primitive action and associated object, and we select the primitive action and associated object with maximum probabilities to achieve the first atomic action, which is fed into the LSTM to predict the second atomic action. This process is repeated until a $(stop, stop)$ atomic action is generated.

\begin{table*}[htbp]
\centering
%\scriptsize
\begin{tabular}
{p{2cm}|p{0.62cm}p{0.62cm}p{0.62cm}p{0.62cm}p{0.62cm}p{0.62cm}p{0.62cm}p{0.62cm}p{0.62cm}p{0.62cm}p{0.62cm}p{0.62cm}p{0.62cm}|p{0.62cm}}
\hline
\multirow{2}{*}{\centering methods} &
\multicolumn{14}{c}{primitive actions} \\
\cline{2-15}
&  move to 	& grasp &	place back	& pour into & open & pour away & hold &	heat & close & turn on & clean & put into & task fail & overall\\
\hline\hline
\centering NN & 88.0 & 70.4 & 36.3 & 36.8 & 54.3 & 61.4 & 71.2 & 30.0 & 39.4 & 42.1 & 93.8 & 43.0 & 32.9 & 63.9 \\
\centering MLP & 98.4 & 93.9 & 86.9 & 87.1 & 92.1 & 100.0 & 96.8 & 88.6 & 83.3 & 100.0 & 93.8 & 87.0 & 90.4  & 93.5\\
\centering RNN & 98.6 & 96.0 & 94.6 & 95.6 & 76.4 & 95.3 & 97.2 & 92.8 & 72.7 & 100.0 & 100.0 & 78.0 & 94.5 &  95.6\\
\hline
\centering Ours w/o AOG & 98.6 & 95.8 & 93.6  & 93.0 & 87.1 & 95.3 & 96.9 & 87.9 & 97.0  & 100.0 & 87.5 & 94.0 &79.5 & 95.4  \\
\centering Ours w/ AOG & 97.9 & 97.0 & 95.6 & 95.7 & 89.3 & 89.8 & 95.4 & 93.8 & 93.9 & 100.0 & 93.8 & 84.0 & 95.9  & \textbf{96.1}  \\
\hline
\hline
\end{tabular}
\begin{tabular}
{p{2cm}|p{0.62cm}p{0.62cm}p{0.62cm}p{0.62cm}p{0.62cm}p{0.62cm}p{0.62cm}p{0.62cm}p{0.62cm}p{0.62cm}p{0.62cm}p{0.62cm}p{0.62cm}|p{0.62cm}}
\hline
\multirow{2}{*}{\centering methods} &
\multicolumn{14}{c}{associated objects} \\
\cline{2-15}
&  cup & pot & water-dis & tea-box & water	 & bowl & easer & board & washing machine & teapot & clothes & closet & task fail & overall\\
\hline\hline
\centering NN & 77.0 & 47.2 & 36.8 & 74.8 & 39.6 & 96.0 & 93.8 & 97.9 & 77.6 & 31.2 & 79.8 & 57.1 &32.9 & 65.2 \\
\centering MLP & 97.8 & 90.6 & 94.4 & 96.2 & 89.7 & 92.0 & 92.7 & 95.8 & 97.8 & 89.6 & 87.9 & 71.4 & 91.8   & 94.2\\
\centering RNN & 97.7 & 94.5 & 100.0 & 96.2  & 91.8 	& 98.7 & 99.0 & 100.0 &93.5 & 92.2  & 86.4  &79.8  & 94.5 & 95.6   \\
\hline
\centering Ours w/o AOG & 97.3 & 94.7 & 94.4 & 95.5 & 90.2 & 100.0 & 95.8  & 95.8 & 97.4 & 98.7 & 98.0 & 90.5 & 79.5 & 95.8  \\
\centering Ours w/ AOG & 97.6 & 96.4 & 88.9 & 92.4 & 93.0 & 100.0 & 95.8 & 95.8 & 100.0 	& 97.4 & 97.0 & 86.9 & 95.9  & \textbf{96.6}  \\
\hline
\end{tabular}
\vspace{4pt}
\caption{Accuracy of the primitive actions and associated objects of our method with and without and-or graph (Ours w/ and w/o AOG, respectively) and the three baseline methods (i.e., RNN, MLP, and NN).}
\label{table:primitive-results}
\end{table*}

%\begin{table*}[htbp]
%\centering
%\scriptsize
%\begin{tabular}
%{p{1.5cm}|p{0.412cm}p{0.412cm}p{0.412cm}p{0.412cm}p{0.412cm}p{0.58cm}p{0.442cm}|p{0.412cm}p{0.412cm}p{0.412cm}p{0.412cm}p{0.412cm}p{0.412cm}p{0.412cm}p{0.412cm}p{0.412cm}p{0.442cm}|p{0.412cm}p{0.412cm}}
%\hline
%\multirow{2}{*}{methods} &
%\multicolumn{13}{c|}{primitive actions} &
%\multicolumn{1}{c}{average} \\
%\cline{2-20}
%& move close to	& grasp &	place back	& pour into & open & pour away & hold &	heat & close & turn on & clean & put into & task fail \\
%\hline\hline
%MLP & 94.50 & 90.70 & 93.20 & 96.60 & 82.20 & 64.30 & 84.80 & 95.20 & 90.90 & 96.30 & 99.20 & 97.70 & 95.90 & 100.00  \\
%RNN & 95.70 & 94.10 & 95.10 & 95.70 & 88.50 & 64.00 & 87.90 & 96.40 & 94.50 & 95.10 & 98.40 & 98.50 & 96.30 & 100.00  \\
%\hline
%Ours w/o AOG & 94.00 & 94.30 & 93.70 & 96.70 & 92.50 & 63.60 & 100.00 & 99.30 & 95.00 & 96.30 & 98.40 & 98.10 & 97.40 \\
%Ours w/ AOG & 95.80 & 95.10 & 94.40 & 96.80 & 94.60 & 63.60 & 100.00 & 99.80 & 95.70 & 96.40 & 98.40 & 98.10 & 98.10 \\
%\hline
%\end{tabular}
%\vspace{2pt}
%\caption{Accuracy results of recognizing atomic actions (i.e., primitive actions and associated objects) generated by our method with and without and-or graph (Ours w/ and w/o AOG, respectively) and the two baseline methods (i.e., RNN and MLP).}
%\label{table:primitive-results}
%\end{table*}

\section{Experiments}
In this section, we introduce the newly collected dataset in detail and present extensive experimental results to demonstrate the superiority of the proposed model. We also conduct experiments to carefully analyze the benefits of the critical components.

\subsection{Dataset construction}
\label{sec:dataset}
To well define the problem of semantic task planning, we create a large dataset that contains 15 daily tasks described by AOGs and 1,284 scene images, with 500 images captured from various scenarios of 7 typical environments, i.e., lab, dormitory, kitchen, office, living room, balcony, and corridor, and the remaining 784 scenes are searched from the Internet, e.g., using Google Image Search. 
All  the objects in these images are annotated with their class labels and initial property. As described above, the atomic action is defined as a two-tuple, i.e., a primitive action and its associated object. In this dataset, we define 12 primitive actions, i.e., ``move to'', ``grasp'', ``place back'', ``pour into'', ``open'', ``pour away'', ``hold'', ``heat'', ``close '', ``turn on '', ``clean'', and ``put into'', and 12 associated objects, i.e., ``cup'', ``pot'', ``water dispenser'', ``tea box'', ``water'', ``bowl'', ``easer'', ``board'', ``washing machine'', ``teapot'', ``clothes'', and ``closet''. Some scenarios exist in which a task cannot be completed. For example, a robot cannot complete the task of ``pour a cup of water from the pot'' if the pot in the scenario is empty. Thus, we further define a specific atomic action $(task fail, task fail)$, and the robot will predict this atomic action when faced with this situation.

The dataset includes three parts, i.e., the training set, the testing set and an augmented set generated from the AOGs. The training set contains 215 samples for 12 tasks with the annotations (i.e., the selections of all the or-nodes in the corresponding AOGs), and this training set is used to train the AOG-LSTM. The augmented set contains 2,600 samples of ${(I, T, \mathcal{A}^p)}$, in which $\mathcal{A}^p$ is the predicted sequence. For training the Action-LSTM, we combine the augmented set and training set. The testing set contains 983 samples of ${(I, T, \mathcal{A})}$ for the performance evaluation. 

\subsection{Experimental settings}

\noindent\textbf{Implementation details. } 
We implement both LSTMs using the Caffe framework \cite{jia2014caffe}, and we train the AOG-LSTM and Action-LSTM using stochastic gradient descent (SGD) \cite{bottou2010large} with a momentum of 0.9, weight decay of 0.0005, batch size of 40, and initial learning rates of 0.1. For curriculum learning, we empirically initialize $\tau$ as 0.2 and add 0.2 to it after training for 100 epochs. The model with the lowest validation error is selected for evaluation.

\noindent\textbf{Evaluation metrics. }We utilize the accuracies of the primitive actions and associated objects, atomic actions, and action sequences as the metrics to evaluate our proposed method. The metrics are described in detail  below. We regard the predicted primitive action as correct if it is exactly the same as the annotated primitive action at the corresponding time step. In addition, the accuracy of the primitive action is defined as the fraction of correctly predicted primitive actions with respect to all primitive actions. The accuracy of the associated object is defined similarly. We regard the predicted atomic action as correct if both the primitive action and its associated object are correct. The accuracy of the atomic action is defined as the fraction of correctly predicted atomic actions with respect to all atomic actions. Finally, we regard the action sequence as correct if the atomic action at each time step is correct. The accuracy of the action sequence is defined as the fraction of correctly predicted sequences with respect to all sequences.

\begin{table}[htp]
\centering
\begin{tabular}{c|c}
\hline
methods & mean acc.\\
\hline\hline
NN & 66.9  \\
MLP & 90.6  \\
RNN & 92.8    \\
\hline
Ours w/o AOG& 93.5  \\
Ours w/ AOG & 96.0   \\
\hline
\end{tabular}
\vspace{2pt}
\caption{Mean accuracy over all atomic actions of our method with and without the and-or graph (Ours w/ and w/o AOG) and the three baseline methods (i.e., RNN, MLP, and NN).}
\label{table:atomic-action}
\end{table}

%\noindent\textcolor[rgb]{1,0,0}{\textbf{Evaluation metrics. } We first evaluate the accuracy of the primitive actions and associated objects individually. Accuracy of primitive action is defined as the fraction of the correctly predicted primitive actions with respect to all primitive actions. Accuracy of associated object is defined in the same way. We also evaluate the accuracy of the atomic action, which can be computed as the fraction of correctly predicted atomic action with respect to all atomic actions. Here, a atomic action is regarded as true if both the primitive action and associated object are both correctly predicted. To comprehensive evaluate the methods, we also present the accuracy of generating the action sequences, i.e., whether the task is completed successfully. It is defined as the fraction of completely correct predicted sequences with respect to all predicted sequences.}

%\begin{figure*}[htbp]
%\centering
%\subfigure[]{
%\includegraphics[width=0.3\linewidth]{action_cm.pdf}}
%\subfigure[]{
%\includegraphics[width=0.3\linewidth]{object_cm.pdf}}
%\caption{The confusion matrices for (a) 7 primitive actions and (b) 10 associated objects of our model.}
%\label{fig:confuse-matrix}
%\end{figure*}

\begin{table*}[htbp]
\centering
%\scriptsize
\begin{tabular}
{p{2.0cm}|p{0.75cm}p{0.75cm}p{0.75cm}p{0.75cm}p{0.75cm}p{0.75cm}p{0.75cm}p{0.75cm}p{0.75cm}p{0.80cm}p{0.80cm}p{0.80cm}|p{0.75cm}}
\hline
methods & task 1  & task 2 & task 3 & task 4 & task 5  & task 6  & task 7 & task 8 & task 9 & task 10  & task 11 & task 12 & overall\\
\hline\hline
\centering NN & 44.0 & 12.9 & 66.7 & 81.2 & 25.4 & 35.7 & 27.9 & 42.0 & 31.2 & 42.1 & 22.5 & 35.7 &33.6\\
\centering MLP & 56.0 & 67.7 & 100.0 & 84.4 & 83.0 & 86.7 & 83.8 & 85.3 & 89.6 & 97.4 & 92.5 & 53.6 & 83.6 \\
\centering RNN & 84.0 & 71.0 & 66.7  & 93.8 & 86.4 	& 90.2 & 86.0 & 86.0 & 92.2 & 97.4 & 70.0 & 25.0 & 84.8 \\
\hline
\centering Ours w/ self aug   & 64.0 & 58.1  & 100.0& 90.6  & 97.2 & 90.9 & 95.5 & 94.4 & 94.8 & 100.0& 92.5 & 67.9 &  91.5 \\
\centering Ours w/o AOG& 92.0 & 80.6 & 100.0 	& 93.8 & 86.9 & 87.4 & 93.3 & 90.9 & 88.3 & 100.0 & 70.0 	& 57.1 & 88.1 \\
\centering Ours w/ AOG & 100.0 & 64.5 & 100.0 &93.8 & 96.6 & 94.4 &93.3 &95.8 & 94.8 & 100.0 & 92.5 &78.6  & \textbf{93.7}\\
\hline
\end{tabular}
\vspace{4pt}
\caption{Sequence accuracy of our method with and without the and-or graph (Ours w/ and w/o AOG), Ours w/ self aug, and the three baseline methods (i.e., RNN, MLP, and NN).  We utilize task 1 to task 12 to denote the ``pour the water in the cup into the bowl'', ``make tea with the cup'', ``make tea with the cup using water from the water dispenser'', ``clean the board'', ``get a cup of hot water'', ``get a cup of hot water from the pot'', ``pour a cup of water'', ``pour a cup of water from the pot'', ``pour a cup of tea from the teapot'', ``wash the clothes with the washing machine'', ``wash the clothes in the washing machine'', and ``put the clothes in the closet'' tasks.}
\label{table:sequence-results}
\end{table*}

% \begin{figure}[htbp]
%    \centering
%    \includegraphics[width=0.8\linewidth]{action_cm.pdf}
%    \caption{The confusion matrices of the primitive actions. Our method can accurately predict the primitive actions.}
%    \label{fig:confuse-matrix-action}
% \end{figure}

% \begin{figure}[htbp]
%    \centering
%    \includegraphics[width=0.8\linewidth]{action_cm.pdf}
%    \caption{The confusion matrices of the associated objects. Our method can accurately predict the associated objects.}
%    \label{fig:confuse-matrix-object}
% \end{figure}

\subsection{Baseline methods}
To verify the effectiveness of our model, we implement three baseline methods that can also be used for semantic task planning for comparison.% For a fair comparison, all the baseline methods are trained on the manually annotated training set.

\subsubsection{Nearest Neighbor (NN)} NN retrieves the most similar scene image and obtains the action sequence that can complete the given task under this image as its output. Concretely, given a new image and task, we extract the image feature and compare it with those on the training set. The sample, which shares the most similar feature with the given image, is taken, and its annotated action sequence regarding the given task is regarded as the final output. The image features are extracted in a similar manner for the AOG-LSTM, as discussed in Section \ref{sec:sagwaog}.

\subsubsection{Multi-Layer Perception (MLP)} We implement an MLP \cite{gardner1998artificial} that predicts the $t$-th atomic action by taking the task features, image features, and previous $t-1$ predicted atomic actions as input. Moreover, it repeats the process until a stop signal is obtained. The MLP consists of two stacked fully connected layers, in which the first layer maps the input to a 512-D feature vector followed by the ReLU function and the second layer maps two vectors followed by softmax layers, which indicate the score distribution of the primitive action and the associated object, respectively. 

\subsubsection{Recurrent Neural Network (RNN)} The training and inference processes and the input features of the RNNs are exactly the same as those of our Action-LSTM. Our method utilizes a traditional hidden state unit rather than an LSTM unit. For a fair comparison, the RNN also has one hidden layer of 512 neurons.

The two baseline methods are also implemented using the Caffe library \cite{jia2014caffe}, and they are trained using SGD with a momentum of 0.9, weight decay of 0.0005, batch size of 40, and initial learning rate of 0.1. We also select the models with the lowest  validation error for a fair comparison. 

\begin{figure}[!t]
\centering
\subfigure[]{
\includegraphics[width=0.48\linewidth]{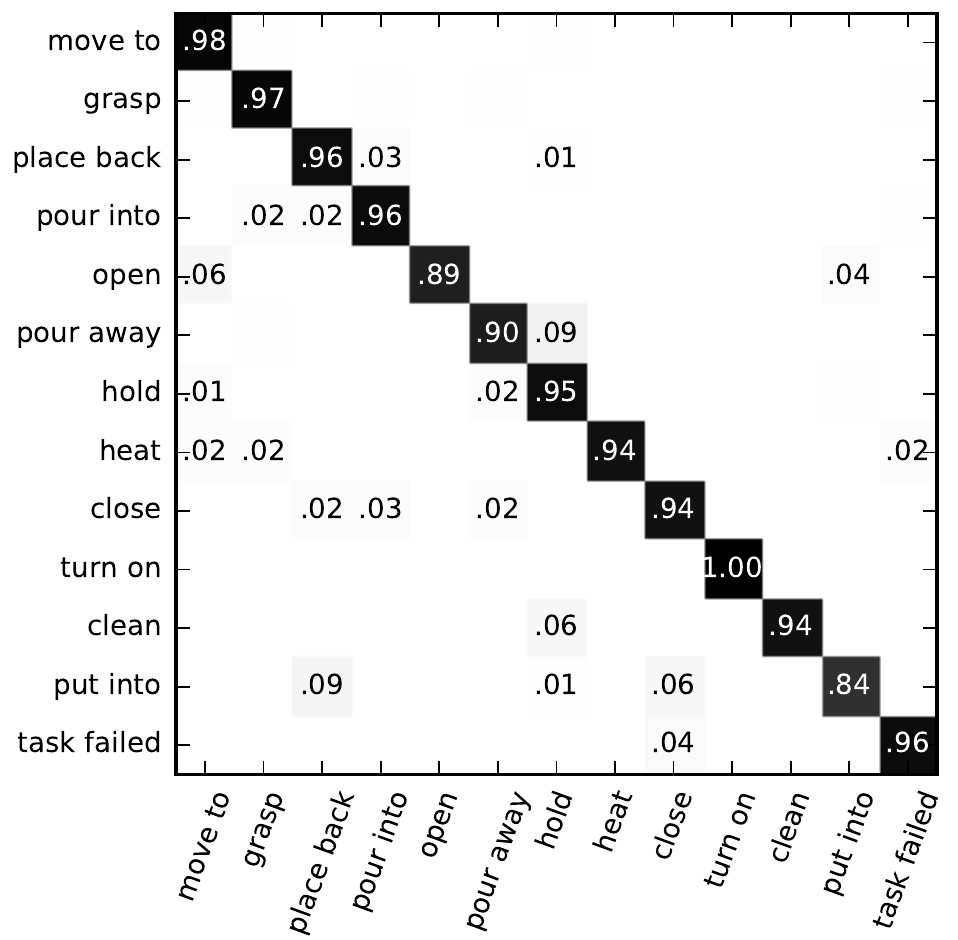}}
\subfigure[]{
\includegraphics[width=0.48\linewidth]{action_cm.pdf}}
\caption{The confusion matrices of the (a) primitive actions and (b) associated objects. Our method can accurately predict both the primitive actions and associated objects.}
\label{fig:confuse-matrix}
\end{figure}

%\subsection{Results and analysis}
\subsection{Comparisons with the baseline models}
We first evaluate the performance of our model for recognizing the primitive actions and associated objects. Figure \ref{fig:confuse-matrix} presents the confusion matrices for these two elements, where our model achieves very high accuracies for most classes. Table \ref{table:primitive-results} further depicts the detailed comparison of our model against the baseline methods. Our model can predict the primitive actions and associated objects with overall accuracies of 96.1\% and 96.6\%, outperforming the baseline methods. We also present the mean accuracy of the atomic action in Table \ref{table:atomic-action}. Here, we compute the accuracy of each atomic action and compute the mean over the accuracies of all the atomic actions. As shown, our model also achieves the highest accuracy compared with the baseline methods.

Then, we evaluate the sequence accuracy of all the methods, as reported in Table \ref{table:sequence-results}. Our model can correctly predict complete action sequences with an overall probability of 93.7\%, evidently outperforming the baseline methods on most tasks (11/12) and improves the overall accuracy by 8.9\%.

\begin{figure}[!t]
   \centering
   \includegraphics[width=1.0\linewidth]{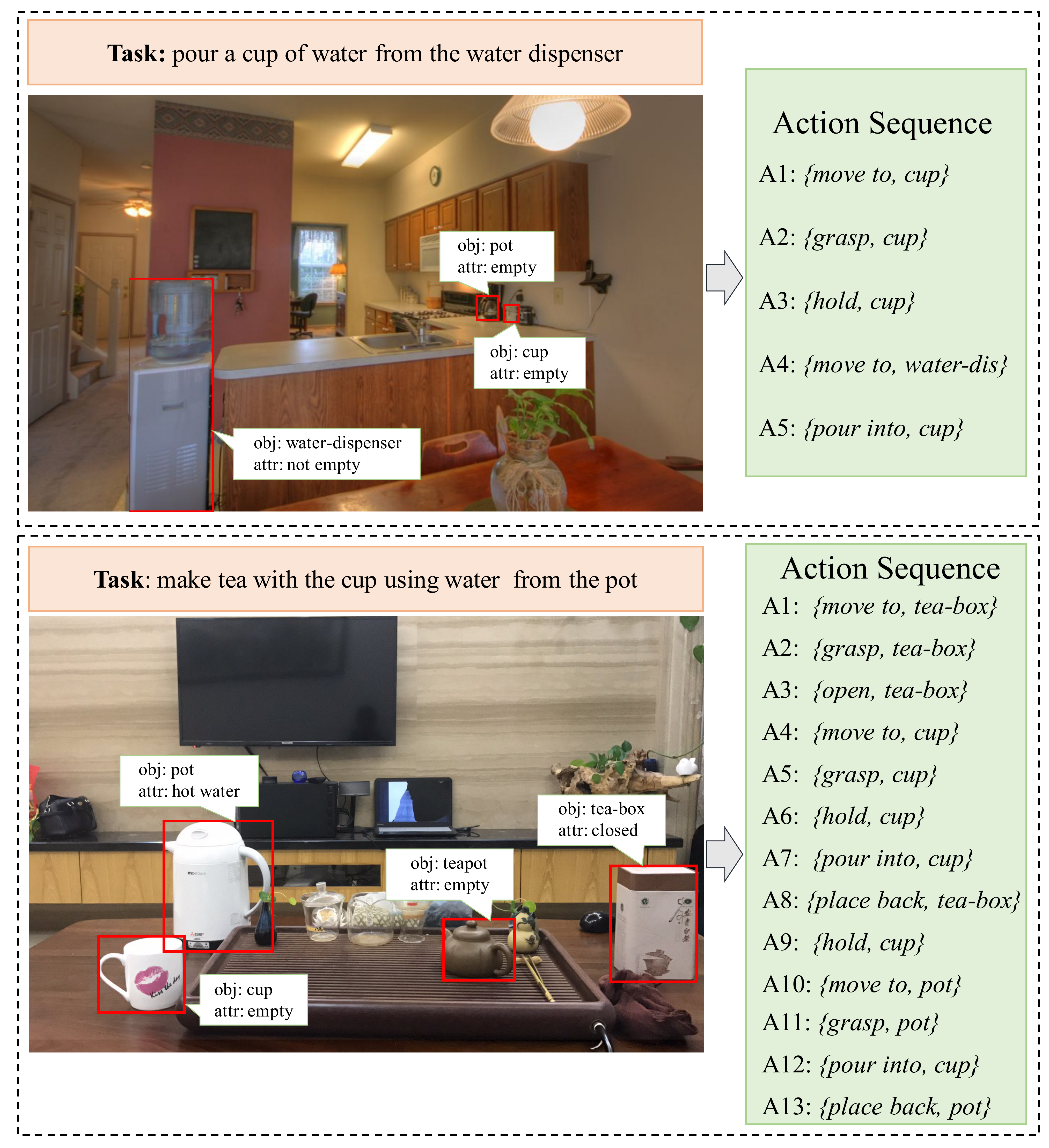}
   \caption{Some atomic action sequences regarding given scene images and tasks generated by our method. Our method is able to correctly predict the action sequence for various tasks across different scenarios.}
   \label{fig:visualization}
\end{figure}

Some atomic action sequences generated by our method are presented in Figure \ref{fig:visualization}. As shown, our method is capable of accurately predicting the action sequences for various tasks across different scenarios. 

\subsection{Generalization to related tasks}
Here, we define ``related tasks'' as tasks that have similar atomic actions or temporal context to the existing tasks in the training set. For example, ``pour a cup of water from the water dispenser'' is a task related to ``pour a cup of water ''. Thus, it would be interesting to see how our trained model can be generalized to  related tasks. In particular, for each related task, we have its AOG representation but no annotated training samples. In this experiment, the models of ``Ours without AOG'' are all trained on the training set, which only contains samples of task 1 to task 12, as described above. For our model with AOG, we first train the AOG-LSTM with the same set of annotated samples as the other competing models. Subsequently, we utilize the trained AOG-LSTM to produce samples for all tasks, including tasks 13, 14 and 15, and then, we use these samples to train the Action-LSTM. The results of the three tasks are presented in Table \ref{table:generlization}. We find that the performances of the method without using AOG are extremely unsatisfying on both tasks.  These results clearly demonstrate the excellent generalization ability of our model, which improves the sequence accuracy by 63.2\%. We also present the sequence accuracy of the AOG-LSTM, i.e., 82.8\%, which is also worse than the proposed model.

% \begin{table}[htp]
% \centering
% \begin{tabular}{c|cccc}
% \hline
% methods & task 13  & task 14 & task 15 & overall\\
% \hline\hline
% MLP & 0.0  & 0.0 & 3.8 & 1.5 \\
% RNN & 0.0  & 64.4  & 61.5 &	48.5  \\
% \hline
% Ours w/o AOG& 0.0 	& 26.9 & 30.8  & 22.1 \\
% Ours w/ AOG & 62.5  &92.3 & 92.3 & 85.3   \\
% \hline
% \end{tabular}
% \vspace{2pt}
% \caption{Sequence accuracy results generated by our model with and without and-or graph (Ours w/ and w/o AOG) and two baseline methods (i.e., RNN and MLP). Tasks 13, 14 and 15 denote tasks ``make tea with the cup using water from the pot'', ``get a cup of hot water from the water dispenser'',  and ``pour a cup of water from the water dispenser'', respectively.}
% \label{table:generlization}
% \end{table}
\begin{table}[htp]
\centering
\begin{tabular}{c|cccc}
\hline
methods & task 13  & task 14 & task 15 & overall\\
\hline\hline
Ours w/ self aug & 6.2& 84.6 & 84.6 & 70.0\\
Ours w/o AOG& 0.0     & 26.9 & 30.8  & 22.1 \\
Ours w/ AOG & 62.5  &92.3 & 92.3 & 85.3   \\
\hline
\end{tabular}
\vspace{2pt}
\caption{Sequence accuracy of our model with and without and-or graph (Ours w/ and w/o AOG) and the Ours w/ self aug. Tasks 13, 14 and 15 denote the ``make tea with the cup using water from the pot'', ``get a cup of hot water from the water dispenser'',  and ``pour a cup of water from the water dispenser'' tasks, respectively.}
\label{table:generlization}
\end{table}

\subsection{Ablation Study}
In this subsection, we perform ablative studies to carefully analyze the contributions of the critical components of
our proposed model.

\subsubsection{Benefit of using and-or graph}
In this experiment, we empirically evaluate the contribution of introducing AOG to the neural network learning. Here, we train the Action-LSTM with and without using the augmented sample set, and we report the results in the last two rows of Table \ref{table:primitive-results} and Table \ref{table:sequence-results}, i.e., Ours w/ and w/o AOG.
It can be observed that the results using AOGs show a notable improvement in both atomic action recognition and sequence prediction. The performance improvements clearly demonstrate the effectiveness of adopting the augmented set. In particular, generating samples from AOG representations enables us to better capture the complex task variations and is an effective way of compensating the neural network learning. Moreover, note that the Action-LSTM network performs better than the traditional RNN model because LSTM is better able to memorize long-term dependencies among actions. 

As discussed above, the trained Action-LSTM can also generate pseudo-labels for unseen samples, and it does not require manually defined AOGs. To see whether the AOGs actually improve the performance, we further implement another baseline (namely, Ours w/ self aug) that uses the Action-LSTM trained on the annotated set to automatically generate a large number of training samples, and then, we train another Action-LSTM using both the annotated and automatically augmented sets. As shown in Table \ref{table:sequence-results}, this achieves an overall sequence accuracy of 91.5\%, better than the baseline Action-LSTM but much worse than our network, i.e., 93.7\%. In addition, when generalizing to unseen tasks, our method has an even more notable improvement over this baseline, i.e., 85.3\% by ours and 70\% by this baseline as shown in Table \ref{table:generlization}.

\subsubsection{Analysis of AOG-LSTM} 
The AOG-LSTM can also generate action sequences by collecting the leaf nodes after all the or-nodes have been selected. Here, we analyze the performance of the AOG-LSTM network. As shown in Table \ref{table:aog}, if both are trained only with the annotated set, the Action-LSTM network performs worse than the AOG-LSTM network. This is because making a selection at the or-nodes is less ambiguous because the AOG representation effectively regularizes the semantic space. Thus, the AOG-LSTM network can achieve a reasonable performance despite being trained using a small number of samples. However, when trained with both the annotated and augmented sets, the Action-LSTM network, in turn, outperforms the AOG-LSTM network. One possible reason is as follows. If giving sufficient training samples, the Action-LSTM network may implicitly learn the semantic structures of the And-Or Graph. Moreover, the Action-LSTM model directly predicts the atomic action, and thus, the predicted atomic action in the previous step may provide strong guidance for the subsequent atomic action prediction. However, the or-node prediction of the AOG-LSTM may not have such a property. Moreover, the Action-LSTM is a more flexible and general framework, and it can also achieve reasonable results without the AOG representation (see Ours w/ AOG). Introducing AOG augmentation can further boost its performance, especially for  unseen tasks (see Table \ref{table:generlization}).

\begin{figure}[htp]
   \centering
   \includegraphics[width=1.0\linewidth]{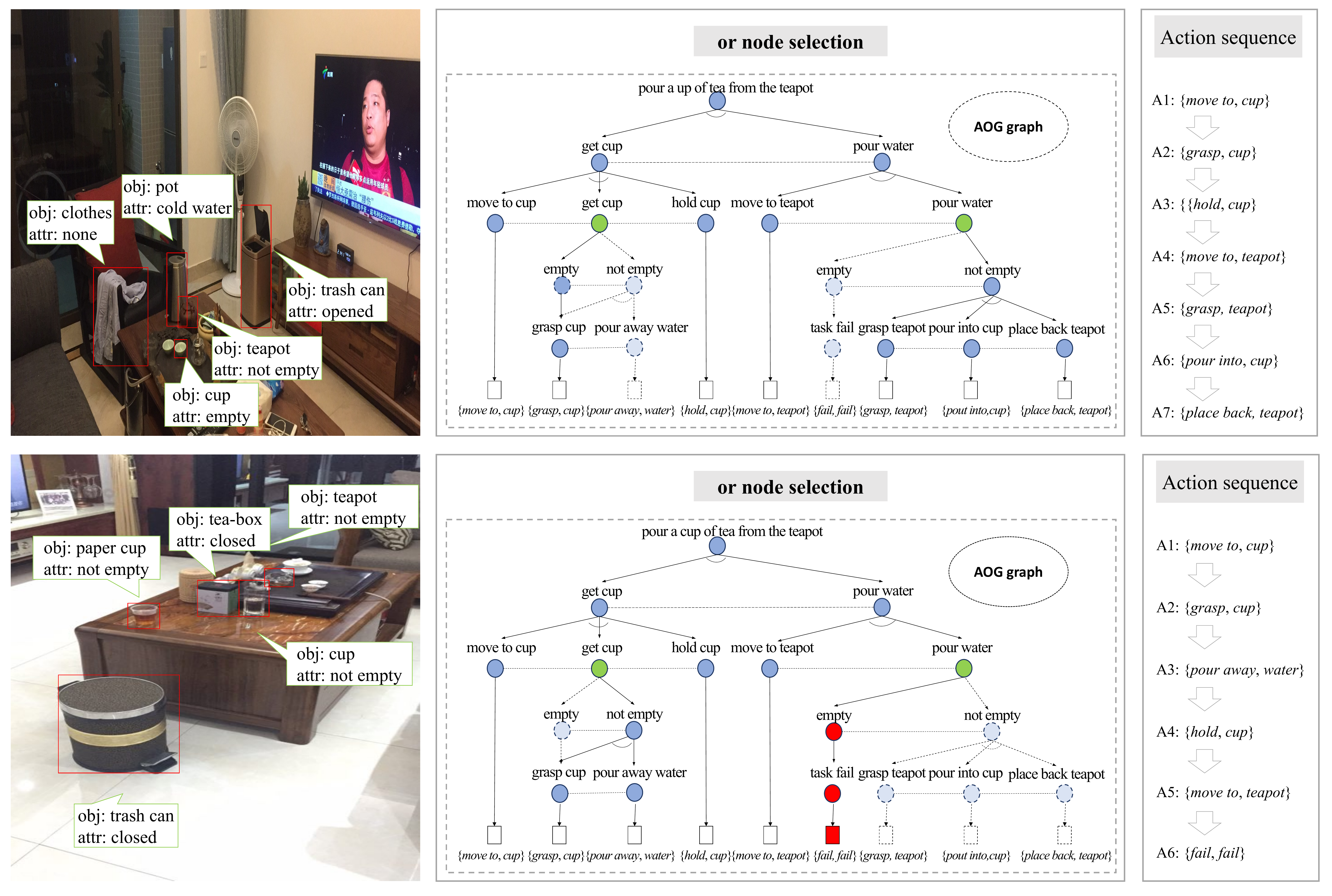}
   \caption{Some samples of or-node selection and the corresponding atomic sequences. The nodes of the unselected branch are denoted as circles with dotted line, and the nodes of incorrectly selected branch are denoted as circles filled with red.}
   \label{fig:aog-visualization}
\end{figure}

Some samples of or-node selection and the corresponding atomic sequences are presented in Figure \ref{fig:aog-visualization}. We find that, for most cases, the AOG-LSTM network can predict the or-node selections correctly, but it is possible to make incorrect predictions if the objects in the image are too complex.

\begin{table}[!t]
\centering
\begin{tabular}{c|c}
\hline
Methods  & Sequence acc.\\
\hline\hline
AOG-LSTM w/o aug  & 91.8 \\
AOG-LSTM w/ aug  & 92.4 \\
Action-LSTM w/o aug  & 88.1 \\
Action-LSTM w/ aug  & 93.7\\
\hline
\end{tabular}
\vspace{2pt}
\caption{Sequence accuracy of the AOG-LSTM network trained with and without augmented set and Action-LSTM trained with and without augmented set.}
\label{table:aog}
\end{table}

\subsubsection{Benefit of curriculum learning}
In this part, we perform an experiment to analyze the contribution of employing the curriculum learning algorithm. Here, we train the Action-LSTM network directly using the entire augmented sample set, and we compare it with our network trained using curriculum learning. The results are reported in Table \ref{table:cl}. As shown, training the network using curriculum learning clearly improves the performance on both atomic action recognition and sequence prediction. This comparison clearly demonstrates the benefit of applying curriculum learning. Concretely, starting the training of the network using the most reliable samples can effectively avoid disturbances incurred by the difficult samples with uncertain or even incorrect labels and thus produce a network with better initialization performance. In this way, we can better utilize the augmented sample set to train the Action-LSTM network. 

%To better understand the training process of curriculum learning, we further illustrate the intermediate results of increasing $\tau$. In particular, we present the results for when we increase $\tau$ by 0.2 every other 10 epochs in Figure \ref{}.%Editor: Please ensure that the intended meaning has been maintained in the above edit. 

\begin{table}[htp]
\centering
\begin{tabular}{c|c}
\hline
Methods  & Sequence acc.\\
\hline\hline
Ours w/o CL  & 92.9 \\
Ours w/ CL  & 93.7\\
\hline
\end{tabular}
\vspace{2pt}
\caption{Sequence accuracy of by our model with and without the curriculum learning (CL) algorithm. Here, we report the sequence accuracy averaged over task 1 to task 12.}
\label{table:cl}
\end{table}

\subsubsection{Benefit of predicting the primitive action and associated object independently}
To address the problem whereby few samples exist for many atomic actions, we simplify the network by assuming the independence of primitive actions and associated objects, and we predict them separately. Here, we conduct an experiment to evaluate the benefit of this simplification. Because there are 35 atomic actions in total, we first remove the two softmax layers in the Action-LSTM network and employ a 35-class softmax layer to directly predict the atomic action, with the other layers left unchanged. We present the sequence accuracy results in Table \ref{table:independency}. Predicting the primitive action and associated object independently can achieve higher sequence accuracies. In particular, this simplification is beneficial for avoiding learning from very few samples and thus enables learning a more robust network.

As discussed in Section \ref{sec:rap}, predicting the primitive action and associated object separately depends on the independence assumption between these two factors. To verify the reasonability of this simplification, we also design some variants that predict the action first and, conditioned on it, predict the object. More concretely, we implement two variants: 1) \emph{Action-LSTM with object condition} shares the same architecture with the proposed Action-LSTM  network except that, at each step, it first predict the score vector of the action and then concatenates it with the hidden state of this step to predict the score vector of the object. 2) \emph{Stacked LSTM with object condition} employs  two stack LSTM networks, in which the first network predicts the score vector of the action, and then, the score vector  together with the hidden state is fed to the second network to predict the score vector of the object. For a fair comparison, we set the dimension of the hidden state as 512 and train the two variants in an identical manner. As shown in Table \ref{table:independency}, the two variants perform slightly worse than the proposed methods. One possible reason for this may be that the prediction of the object also depends on the predicted action, and these dependencies are also rare in the training set. These comparisons show that this simplification can simplify the network while also improving performance.

\begin{table}[htp]
\centering
\begin{tabular}{c|c}
\hline
Methods & Sequence acc.\\
\hline\hline
Action-LSTM (joint)  & 91.3\\
Action-LSTM (condition)& 92.3 \\
Stacked LSTM (condition)& 92.6\\
Action-LSTM & 93.7\\
\hline
\end{tabular}
\vspace{2pt}
\caption{Sequence accuracy of Action-LSTM predicting primitive action and associated object separately (Action-LSTM), directly predicting the atomic action (Action-LSTM (joint)), Action-LSTM with object condition (Action-LSTM (condition)) and Stacked LSTM with object condition (Stacked LSTM (condition)). Here, we report the sequence accuracy averaged over task 1 to task 12.}
\label{table:independency}
\end{table}

\subsubsection{Evaluation of task embedding}
In this work, we use the one-hot vector for task encoding because there are only 15 tasks, and this simple method can well represent each task. To compare this method with other embedding methods, we also conduct an experiment that utilizes semantic embedding for the tasks. Concretely, we use the trained GloVe model \cite{pennington2014glove} to encode a semantic vector for each word of a specific task and average the vectors of all words to achieve the representation of this task. We use this representation to replace the one-hot encoding and re-train the AOG-LSTM and Action-LSTM. We find that the overall sequence accuracy drops from 93.7\% to 92.3\%, as shown in Table \ref{table:embedding}. One possible reason for this may be as follows. There are only 15 tasks;  simple one-hot encoding can well represent each task. If using the more complex semantic representation, by learning from merely 15 sentences, it may be difficult to capture the differences among different tasks.

\begin{table}[htp]
\centering
\begin{tabular}{c|c}
\hline
Methods & Sequence acc.\\
\hline\hline
Ours (GloVe)  & 92.3\\
Ours (one-hot) & 93.7\\
\hline
\end{tabular}
\vspace{2pt}
\caption{Sequence accuracy of our method using Glove and one-hot encoding for  task embedding. Here, we report the sequence accuracy averaged over task 1 to task 12.}
\label{table:embedding}
\end{table}

% \centering
% \begin{tabular}{c|c}
% \hline
% methods & sequence acc.\\
% \hline\hline
% Ours (GloVe)  & 92.3\\
% Ours (one-hot) & 93.7\\
% \hline
% \end{tabular}
% \vspace{2pt}
% \caption{Sequence accuracy results generated by our method using Glove and one-hot encoding for the task embedding. Here, we report the sequence accuracy averaged over task 1 to task 12.}
% \label{table:embedding}
% \end{table}

\subsection{Results for noisy environments}
The above-mentioned experiments are conducted under the assumption of perfect object/attribute detection. It is more practical to evaluate the method in noisy environments. To this end, we further conduct an experiment on noise settings. Specifically, we add Gaussian noise to the one-hot vector of the class label and those of the attributes, and thus, the one-hot vector becomes the score vector, with each element denoting the confidence of the corresponding category or state. We regard the score vector as positive if the bit corresponding to the ground-truth labels has the largest value; otherwise, it is regarded as negative. We add different levels of noise to obtain different negative ratios (e.g., 10\% and 20\%) in both the training and test set, and we re-train the Action-LSTM network. As shown in Table \ref{table:noise}, our network with negative ratios of 10\% and 20\% error labels achieves sequence accuracies of 46.2\% and 41.2\%, respectively.

To better evaluate the performance in a real vision system, we further train a Faster R-CNN detector \cite{ren2015faster} on the training set to automatically detect objects in the given image. We still train the AOG-LSTM and Action-LSTM networks using the annotated objects and evaluate on the test set using the objects detected by the detector. As shown in Table \ref{table:noise}, our network can also achieve reasonable results, e.g., an overall sequence accuracy of 55.0\%.

\begin{table}[htp]
\centering
\begin{tabular}{c|c}
\hline
Methods & Sequence acc.\\
\hline\hline
Ours w/ 10\% noise   & 46.6 \\
Ours w/ 20\% noise & 41.2\\
Ours using detector & 55.0\\
Ours w/o noise  & 93.7\\
\hline
\end{tabular}
\vspace{2pt}
\caption{Sequence accuracy of our model under the setting of perfect detection, 10\%noise and 20\% noise (Ours w/o noise, Ours w/ 10\% noise,  and Ours w/ 20\% noise), and automatic detection (Ours using detector). Here, we report the sequence accuracy averaged over task 1 to task 12, where the networks are only trained with the manually annotated set.}
\label{table:noise}
\end{table}

\vspace{6pt}

\section{Conclusion}
In this paper, we address a challenging problem, i.e., predicting a sequence of actions to accomplish a specific task under a certain scene, by developing a recurrent LSTM network. To alleviate the issue of requiring large amounts of annotated data, we present a two-stage model training approach by employing a knowledge AOG representation. From this representation, we can produce a large number of valid samples (i.e., task-oriented action sequences) that facilitate learning of the LSTM network. Extensive experiments on a newly created dataset demonstrate the effectiveness and flexibility of our approach. 

This is an early attempt to address the task of semantic task planning, but there are certain limitations that prevent the proposed method from extending to more realistic setups. First, the images are pre-processed into a handcrafted feature vector that contains  information about the object categories and locations. This pre-processing prevents the model from using end-to-end training and being robust to inference, and these low-dimensional features can only capture limited characteristics of the visual scene. %Editor: Please ensure that the intended meaning has been maintained in the above edit.
Second, the model was only evaluated on still images; it is unclear if it can easily be extended to  a real robot to perform tasks. Third, the structure of the AOG is manually defined; this can be expensive to collect and is a less flexible option. In future work, we will resort to  simulation platforms such as AI2-THOR \cite{zhu2017target,kolve2017ai2} to collect large numbers of annotated samples to train the detectors and classifiers. In this way, we can extricate the model from dependencies on handcrafted image pre-processing, automatically detect objects in a scene and predict their initial states of the attributes. Moreover, we can also enable an agent to interact with objects and perform tasks on these platforms to evaluate our model. On the other hand, we will also explore automatically learning the AOG structure from annotated samples, thereby improving the flexibility and extendibility of the proposed method.

{
\bibliographystyle{IEEEtran}
\bibliography{reference}
}

\begin{IEEEbiography}[{\includegraphics[width=1in,height=1.25in,clip,keepaspectratio]{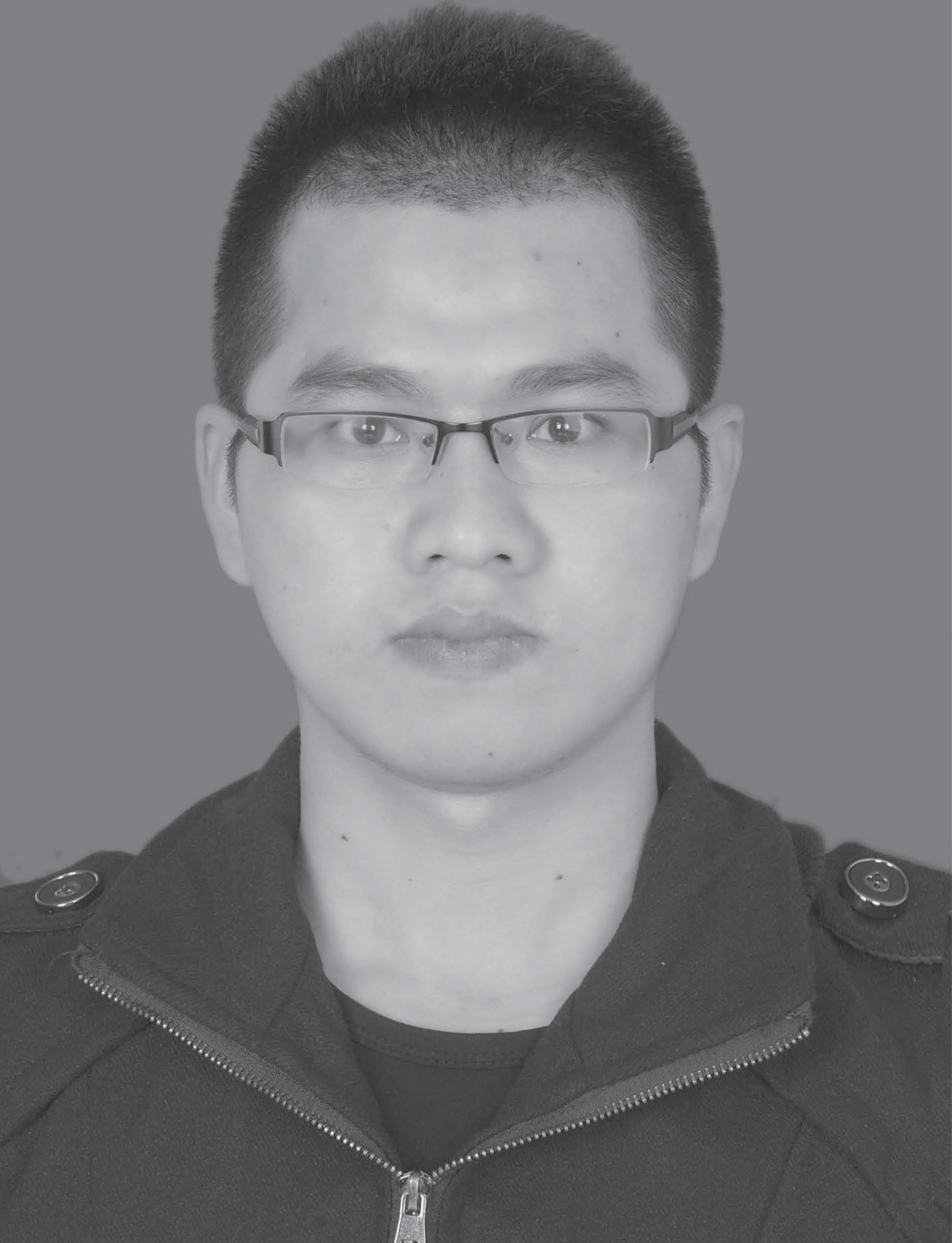}}]{Tianshui Chen} 
received his B.E. degree from the School of Information and Science Technology, Sun Yat-sen University, Guangzhou, China, in 2013. He is currently pursuing his Ph.D. degree in computer science with the School of Data and Computer Science. His current research interests include computer vision and machine learning. He was the recipient of Best Paper Diamond Award in IEEE ICME 2017.
\end{IEEEbiography}

\begin{IEEEbiography}[{\includegraphics[width=1in,height=1.25in,clip,keepaspectratio]{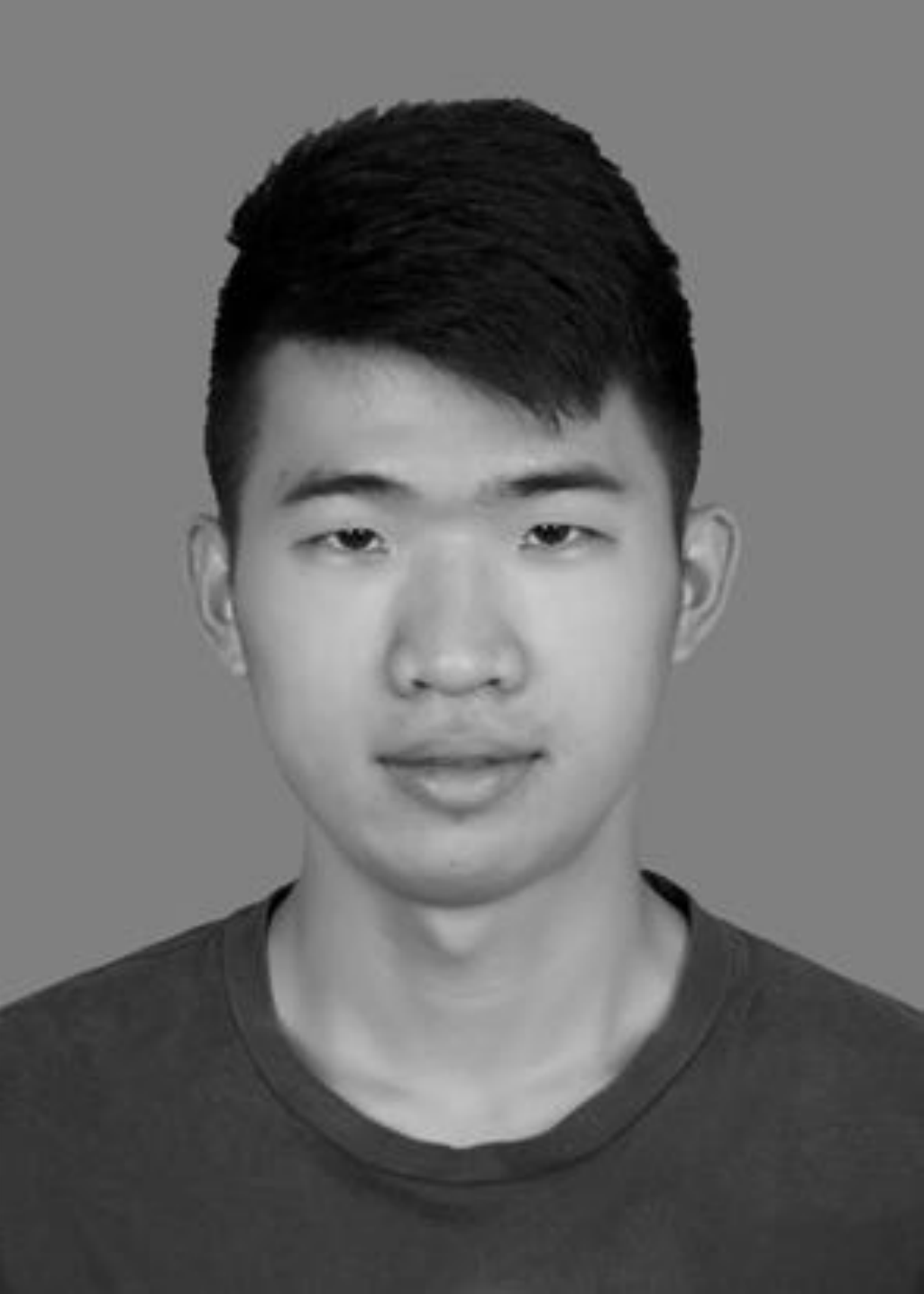}}]{Riquan Chen} 
received his B.E. degree from the School of Mathematics, Sun Yat-sen University, Guangzhou, China, in 2017, where he is currently pursuing his Master's Degree in computer science with the School of Data and Computer Science. His current research interests include computer vision and machine learning.
\end{IEEEbiography}

\begin{IEEEbiography}[{\includegraphics[width=1in,height=1.25in,clip,keepaspectratio]{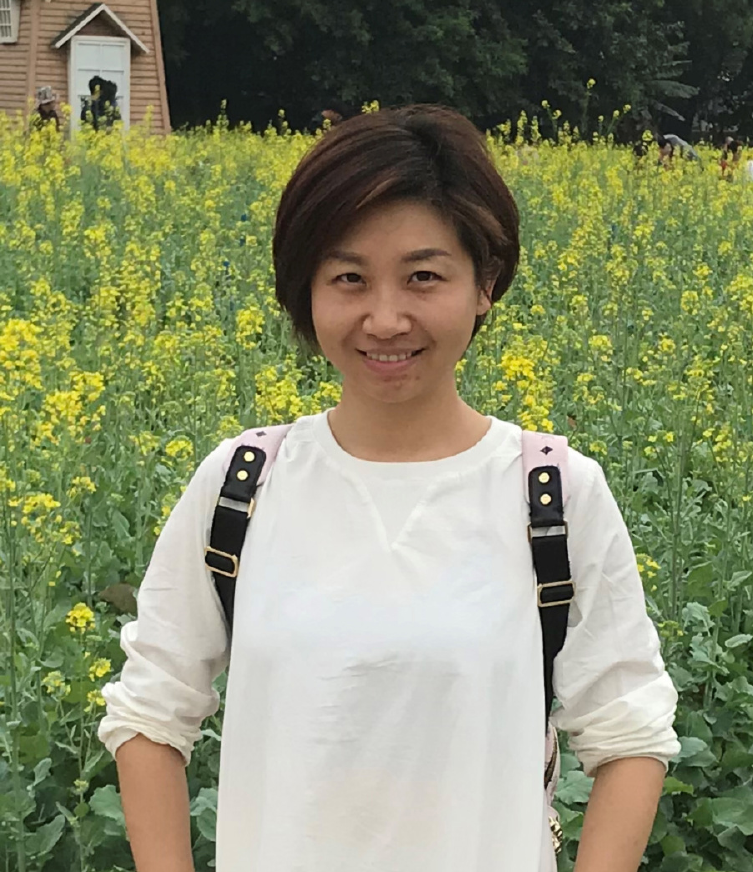}}]{Lin Nie} is an Assistant Researcher with the Center for Shared Experimental Education, Sun Yat-Sen University (SYSU), China. She received her B.S. degree from the Beijing Institute of Technology (BIT), China in 2004 and her M.S. degree from the Department of Statistics, University of California, Los Angeles (UCLA). Her research focuses on data mining and pattern recognition.
\end{IEEEbiography}

\begin{IEEEbiography}[{\includegraphics[width=1in,height=1.25in,clip,keepaspectratio]{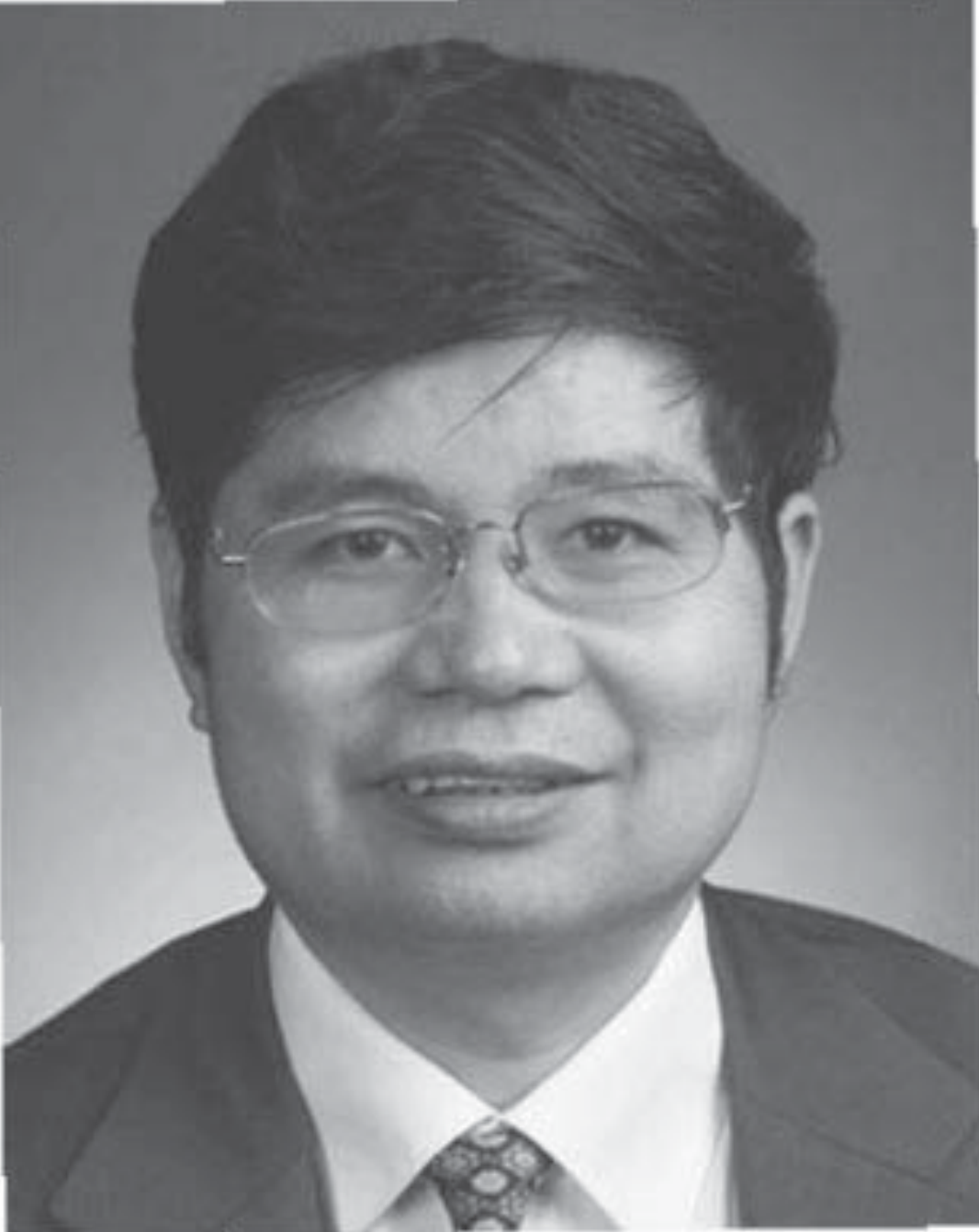}}]{Xiaonan Luo} was the Director of the National Engineering Research Center of Digital Life, Sun Yat-sen University, Guangzhou, China. He is currently a Professor with the School of Computer Science and Information Security, Guilin University of Electronic Technology, Guilin, China. He received the National Science Fund for Distinguished Young Scholars granted by the National Nature Science Foundation of China. His research interests include computer graphics, CAD, image processing, and mobile computing.
\end{IEEEbiography}

\begin{IEEEbiography}[{\includegraphics[width=1in,height=1.25in,clip,keepaspectratio]{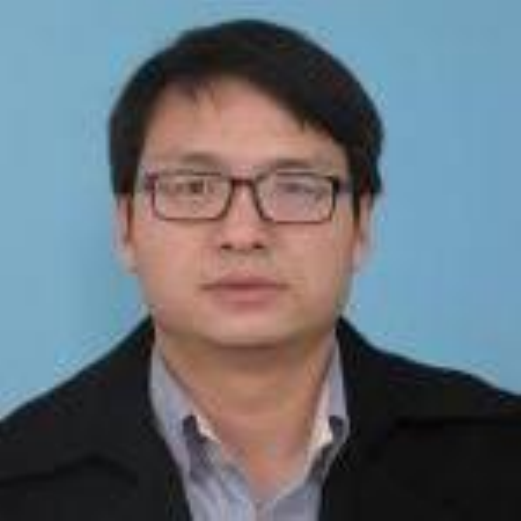}}]{Xiaobai Liu} is an Assistant Professor of Computer Science at San Diego State University (SDSU), San Diego, U.S.A. He received his PhD from the Huazhong University of Science and Technology (HUST), China. His research interests focus on the development of theories, algorithms, and models for core computer vision problems, including image parsing, 3D vision, and visual tracking. He has published 50 peer-reviewed articles in top-tier conferences (e.g.,ICCV and CVPR) and leading journals (e.g., TPAMI and TIP). He received a number of awards for his academic contributions, including the outstanding thesis award by CCF (China Computer Federation).
\end{IEEEbiography}

\begin{IEEEbiography}[{\includegraphics[width=1in,clip]{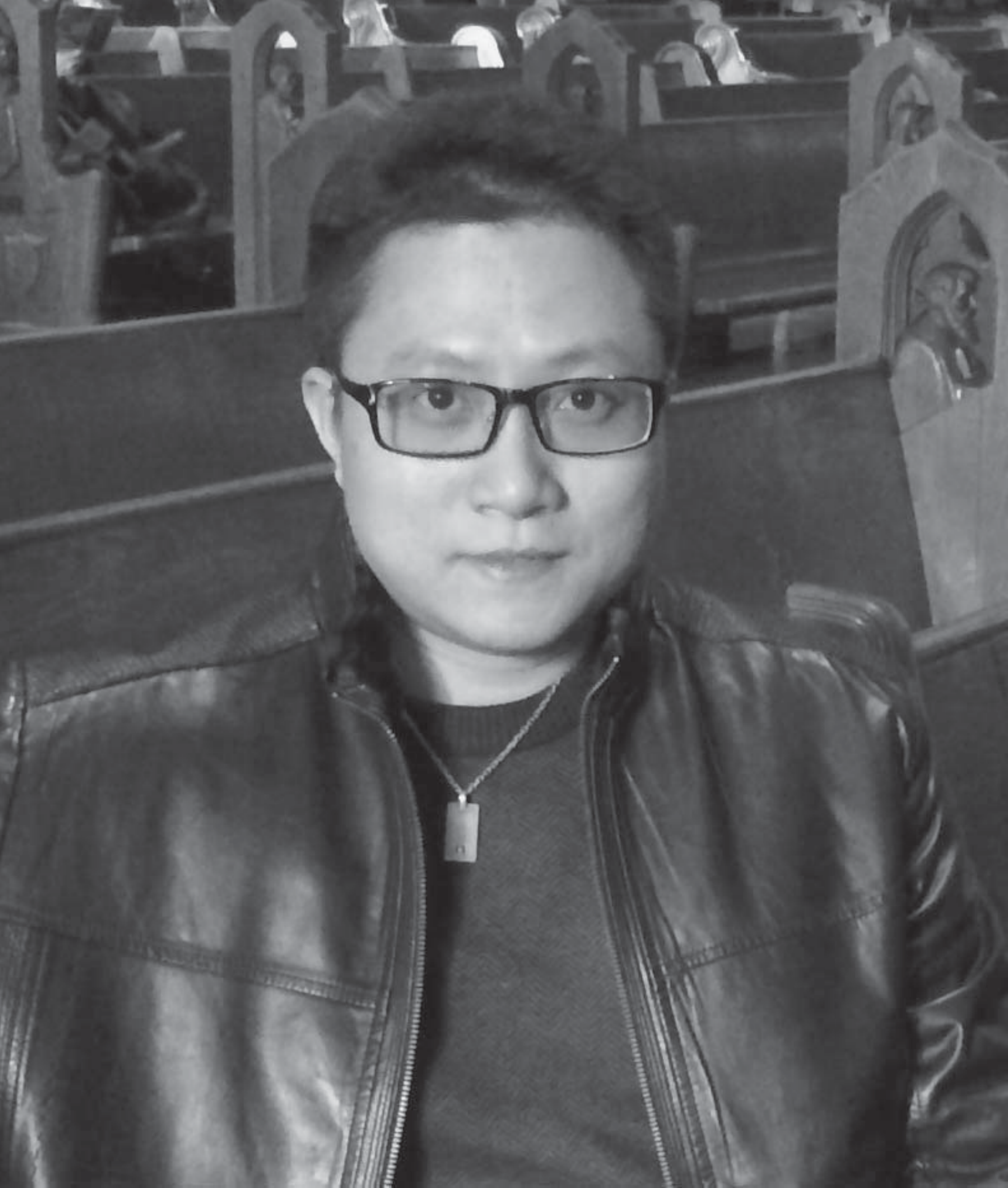}}]{Liang Lin} (M'09, SM'15) is the Executive Research Director of SenseTime Group Limited and a full Professor of Sun Yat-sen University. He is the Excellent Young Scientist of the National Natural Science Foundation of China. From 2008 to 2010, he was a Post-Doctoral Fellow at University of California, Los Angeles. From 2014 to 2015, as a senior visiting scholar, he was with The Hong Kong Polytechnic University and The Chinese University of Hong Kong. He currently leads the SenseTime R\&D teams in developing cutting-edge and deliverable solutions on computer vision, data analysis and mining, and intelligent robotic systems. He has authored and co-authored more than 100 papers in top-tier academic journals and conferences (e.g., 10 papers in TPAMI/IJCV and 40+ papers in CVPR/ICCV/NIPS/IJCAI). He has been serving as an associate editor of IEEE Trans. Human-Machine Systems, The Visual Computer and Neurocomputing. He served as Area/Session Chair for numerous conferences such as ICME, ACCV, and ICMR. He was the recipient of Best Paper Runner-Up Award in ACM NPAR 2010, the Google Faculty Award in 2012, the Best Paper Diamond Award at IEEE ICME 2017, and Hong Kong Scholars Award in 2014.  He is a Fellow of IET.
\end{IEEEbiography}

\end{document}